\documentclass[11pt]{article}

\usepackage[final]{acl}

\usepackage{times}
\usepackage{latexsym}
\usepackage[most]{tcolorbox}
\usepackage[T1]{fontenc}
\usepackage{amssymb}
\usepackage{amsmath}
\usepackage{float}
\usepackage{booktabs}
\usepackage{algorithm}
\usepackage{algpseudocode}
\usepackage{enumitem}
\usepackage{listings}
\usepackage{xcolor}

\usepackage[utf8]{inputenc}

\usepackage{microtype}
\usepackage{multirow}

\usepackage{inconsolata}

\usepackage{graphicx}

\title{DisastQA: A Comprehensive Benchmark for Evaluating Question Answering in Disaster Management}

\author{
\textbf{Zhitong Chen}\textsuperscript{*}\quad
\textbf{Kai Yin}\textsuperscript{*}\quad
\textbf{Xiangjue Dong}\quad
\textbf{Chengkai Liu}\quad
\textbf{Xiangpeng Li}\quad
\textbf{Yiming Xiao}\\
\textbf{Bo Li}\textsuperscript{\ensuremath{\dagger}}\quad
\textbf{Junwei Ma}\textsuperscript{\ensuremath{\dagger}}\quad
\textbf{Ali Mostafavi}\quad
\textbf{James Caverlee}\\
Texas A\&M University\\
\texttt{\{zhitong.chen18,kai.yin,xj.dong,liuchengkai,xplli\}@tamu.edu}\\
\texttt{\{libo,jwma,yxiao,mostafavi,caverlee\}@tamu.edu}
}

\begin{document}
\maketitle
\begingroup
\renewcommand\thefootnote{\fnsymbol{footnote}}
\footnotetext[1]{Equal contribution.}
\footnotetext[2]{Corresponding authors.}
\endgroup

\begin{abstract}
Accurate question answering (QA) in \textbf{disaster management} requires reasoning over uncertain and conflicting information, a setting poorly captured by existing benchmarks built on clean evidence.
We introduce \textsc{DisastQA}, a large-scale benchmark of 3,000 rigorously verified questions (2,000 multiple-choice and 1,000 open-ended) spanning eight disaster types.
The benchmark is constructed via a human--LLM collaboration pipeline with stratified sampling to ensure balanced coverage.
Models are evaluated under varying evidence conditions, from closed-book to noisy evidence integration, enabling separation of internal knowledge from reasoning under imperfect information.
For open-ended QA, we propose a human-verified keypoint-based evaluation protocol emphasizing factual completeness over verbosity.
Experiments with 20 models reveal substantial divergences from general-purpose leaderboards such as MMLU-Pro.
While recent open-weight models approach proprietary systems in clean settings, performance degrades sharply under realistic noise, exposing critical reliability gaps for disaster response.
All code, data, and evaluation resources are available at \href{https://github.com/TamuChen18/DisastQA_open}{the project page}.
\end{abstract}

\section{Introduction}
\label{sec:intro}

Disasters inevitably trigger severe societal, economic, and humanitarian consequences. 
Reliable question answering (QA) systems crucially support timely decision-making, situational awareness, and coordinated responses during crises~\citep{palen2016crisis,vieweg2010microblogging}. 
Large Language Models (LLMs) hold great promise for advancing such capabilities by leveraging both their \textit{parametric knowledge} and \textit{external evidence}.
However, with the rapid evolution of LLMs, a key question arises: \textit{can even the latest frontier models maintain reliability in these high-stakes scenarios?}

Disaster-response QA differs fundamentally from general-domain QA. 
Information in this domain is highly fragmented across heterogeneous sources, ranging from official bulletins and scientific reports to unstructured social media streams~\citep{imran2015processing,alam2021crisisbench}. 
These sources are often incomplete, noisy, or contradictory, and answering disaster-related questions requires synthesizing multiple factual aspects such as hazard type, affected regions, and resource availability. 
This makes \textit{factual completeness and accurate information integration} essential for reliability.

\begin{figure}[t]
    \centering
    \includegraphics[width=\columnwidth]{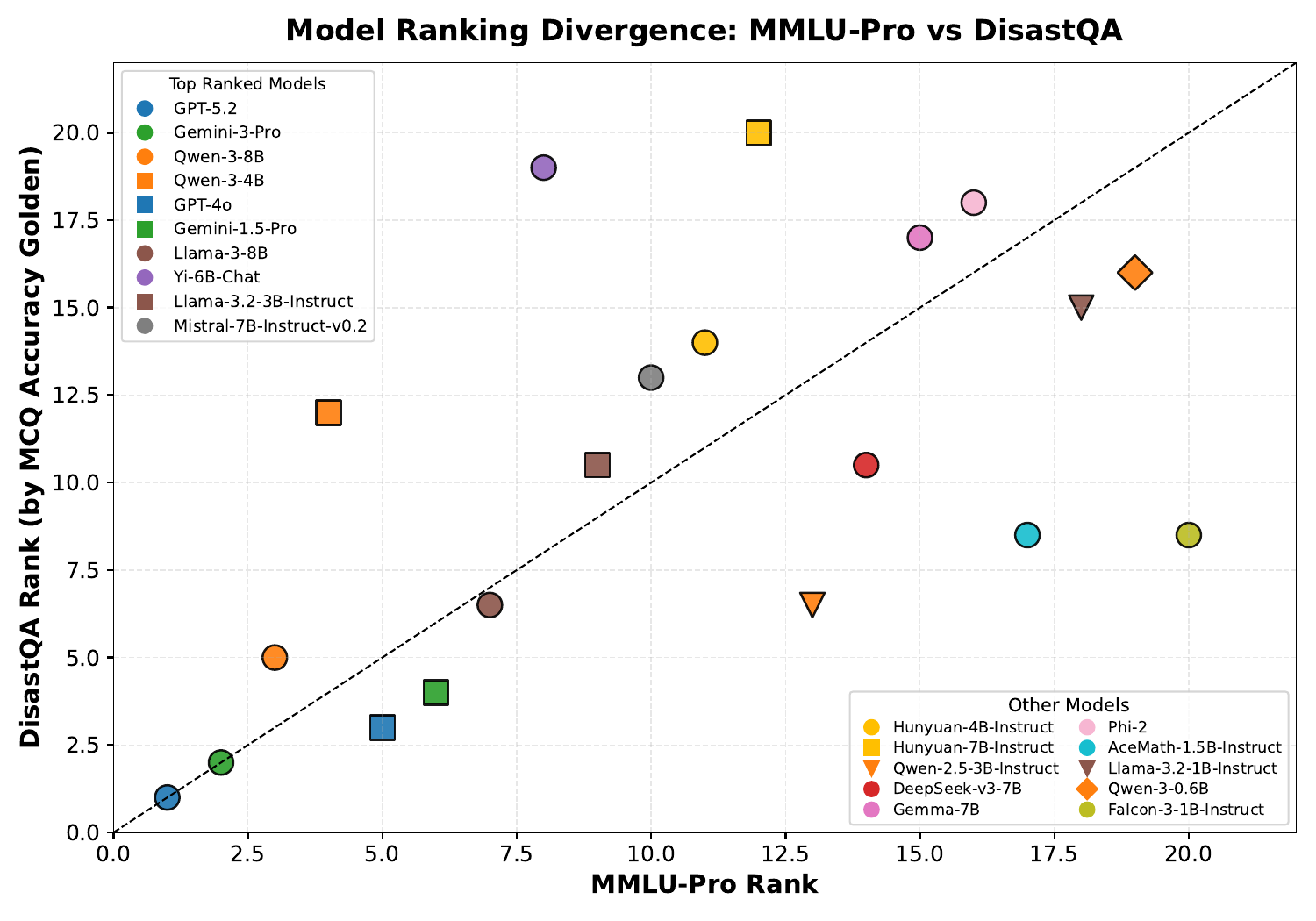}
    \caption{
Model ranking divergence between DisastQA-MCQ (Gold) and the MMLU-Pro subset.
Pronounced off-diagonal deviations indicate that general QA leaderboards may not reliably predict relative performance in high-stakes disaster-response QA.
Full numerical results are provided in Table~\ref{tab:mmlu_subset}.
}
    \label{fig:mcq_performance}
\end{figure}

\begin{figure*}[t]
    \centering
    \includegraphics[width=\textwidth]{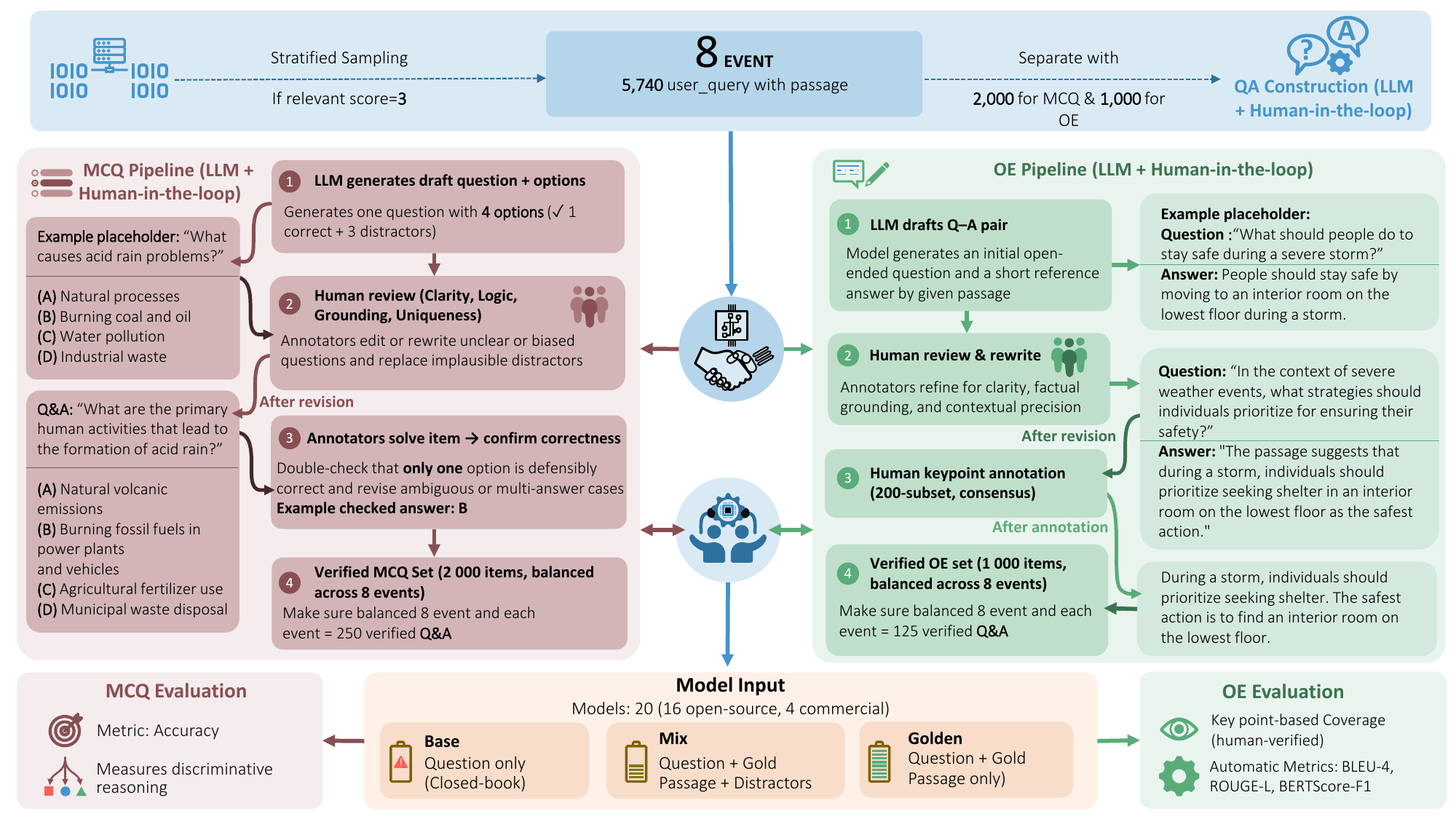}
\caption{The \textbf{Human--LLM Collaborative Pipeline} for DisastQA.
User queries are stratified across 8 disaster event types (\textbf{Top}).
Two parallel tracks construct the dataset: the \textbf{MCQ Pipeline} with human refinement for distractor quality and answer correctness, and the \textbf{OE Pipeline} with human rewriting and \textbf{Keypoint} annotation (\textbf{Middle}).
Models are evaluated under three evidence settings (\textit{Base}, \textit{Mix}, \textit{Golden}) (\textbf{Bottom}); see Section~\ref{sec:evaluation_methodology} for details.}
\label{fig:pipeline}

    \label{fig:pipeline}
\end{figure*}

However, existing QA and evaluation benchmarks mainly target general knowledge~\citep{rajpurkar2016squad,joshi2017triviaqa,hendrycks2021measuring} 
or domains such as medicine and law~\citep{jin2019pubmedqa,zhong2024agieval}, 
leaving disaster contexts underexplored.
Furthermore, most existing datasets are \textit{multiple-choice (MCQ)} based, failing to assess reasoning depth and factual completeness. 
In disaster management, decision-makers often require \textit{open-ended, evidence-grounded responses}, e.g., describing affected regions, resource needs, or response measures, rather than selecting from fixed options.
Thus, existing QA benchmarks fall short in evaluating reliability and completeness under decision-critical conditions.

To address these gaps, we introduce DisastQA, the first large-scale benchmark for disaster-response QA integrating both Multiple-Choice (MCQ) and Open-Ended (OE) tasks. 
Developing such a benchmark poses two main challenges: 
(1) evaluating open-ended responses in an interpretable and reliable way that captures factual completeness, and 
(2) constructing large-scale, high-quality QA data without prohibitive manual annotation costs.

To tackle these challenges, DisastQA adopts a \textbf{Human--LLM collaboration pipeline} (Figure~\ref{fig:pipeline}) that balances scalability with rigorous human oversight, ensuring high-quality, factually grounded, and diverse QA pairs across eight major disaster categories.
We explicitly evaluate models under distinct information settings—ranging from closed-book generation to reasoning with noisy retrieval results.
This design disentangles a model's internal knowledge from its ability to reason with external information.
For open-ended QA, we introduce \textbf{Keypoint Coverage}, a metric designed to measure factual completeness beyond lexical overlap, addressing the limitations of surface-level metrics such as ROUGE and BLEU.
Together, these designs make DisastQA a systematic framework for evaluating both knowledge-based and evidence-grounded capabilities of LLMs.

Using \textsc{DisastQA}, we evaluate 20 open-source and commercial LLMs, including frontier systems such as GPT-5.2 and Gemini-3 Pro.
We compare against a general-domain benchmark (MMLU-Pro) and assess factual coverage in OE tasks.
Our results reveal two critical insights:
(1) Closing Performance Gaps: Efficient open-weight models such as Qwen-3-8B now approach proprietary leaders (e.g., GPT-4o), indicating a narrowing capability gap in this domain.
(2) The Persistence of Reasoning Gaps: Even the strongest frontier models degrade under retrieval noise (i.e., top-$k$ retrieved contexts containing both relevant evidence and plausible distractor passages) and fail to achieve perfect Keypoint Coverage in complex scenarios.
These findings highlight that while scaling improves general capabilities, domain-specific reliability in safety-critical contexts remains an unsolved challenge.

Our Contributions.
We (1) construct the first large-scale, human-verified benchmark for disaster QA through a Human-LLM collaboration pipeline; 
(2) introduce a transferable keypoint-based evaluation framework to measure factual completeness in open-ended QA; and 
(3) conduct comprehensive evaluations of 20 LLMs, establishing new upper bounds with frontier models while revealing persistent reliability gaps in disaster-specific reasoning.

\section{Related Work} \label{sec:related_work}
\paragraph{Existing QA Benchmarks.}
Standard QA benchmarks emphasize broad coverage but \textbf{overlook} the reliability constraints of disaster-response settings.
For example, SQuAD-style reading-comprehension datasets~\citep{rajpurkar2016squad} \textbf{assume relatively clean}, single-context evidence, missing the uncertainty common during crises.
Large-scale long-form QA datasets such as ELI5~\cite{fan-etal-2019-eli5} and GOOAQ~\cite{khashabi-etal-2021-gooaq} rely on web documents, while WikiCQA~\cite{dong-etal-2023-closed} targets closed-book settings.
Meanwhile, benchmarks such as MMLU-Pro~\citep{wang2024mmlu-pro} mainly probe parametric knowledge through multiple-choice questions, \textbf{limiting evaluation of} evidence-grounded, open-ended synthesis.
As a result, neither paradigm directly tests robustness to retrieval noise or the \textit{factual completeness} required for decision support.

\paragraph{Disaster-Focused Resources.} 
Prior work in disaster informatics has focused on information extraction tasks such as classification, retrieval, and summarization~\citep{imran2015processing, olteanu2014crisislex, rudra2019summarizing}, rather than generative QA. 
While a few attempts like DisasterQA~\citep{rawat2024disasterqa} exist, they are constrained by small scale, reliance on multiple-choice formats, and a lack of controlled uncertainty evaluation. 
\textbf{DisastQA} bridges these gaps by integrating both Multiple-Choice and Open-Ended tasks under a tri-level evaluation setup (\textit{Base}, \textit{Mix}, \textit{Golden}) to disentangle internal knowledge from evidence reasoning. 
Furthermore, we introduce a \textit{keypoint-based protocol} to rigorously quantify factual completeness, offering the first large-scale, reproducible, and evidence-grounded benchmark for the domain.

\begin{figure}[t]
\centering
\includegraphics[width=\columnwidth]{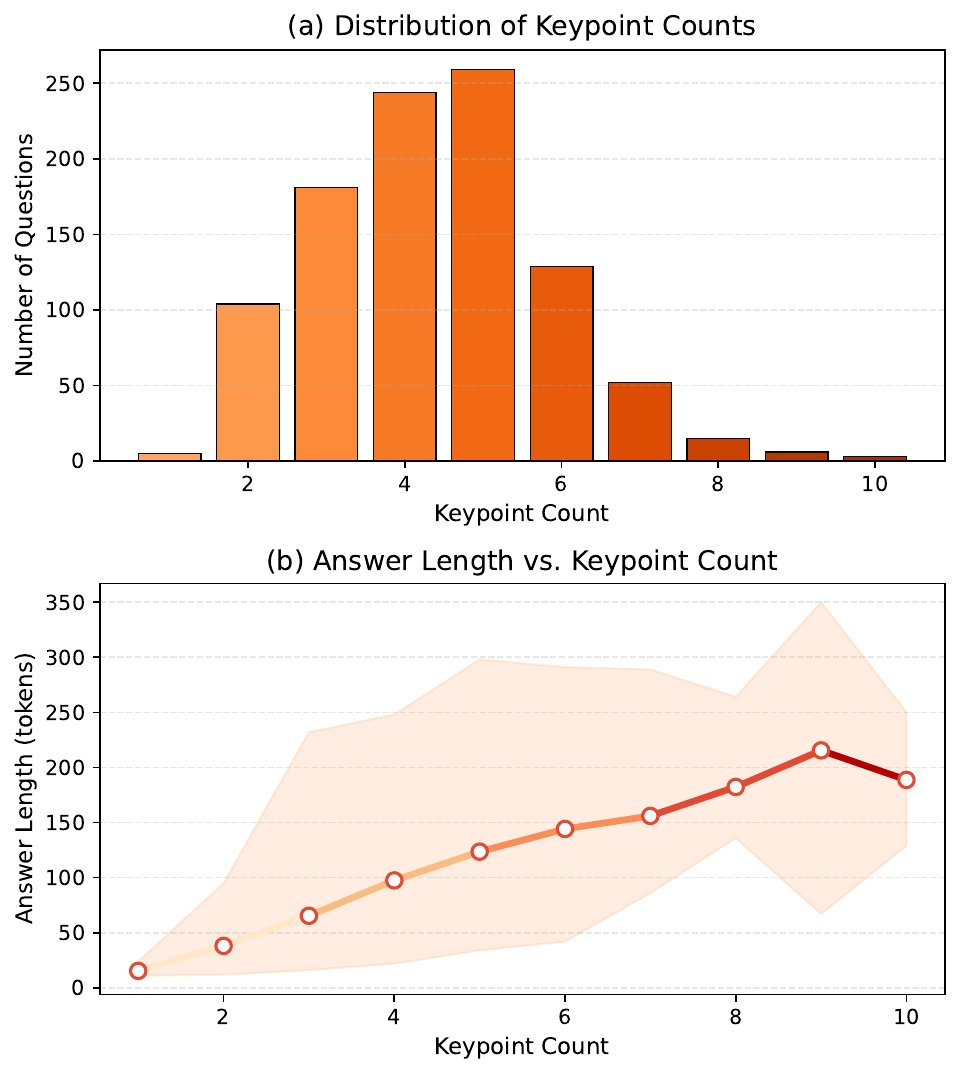}
\caption{
(a) Distribution of keypoint counts per OE answer. 
(b) Relationship between answer length (in tokens) and keypoint count, showing that longer answers typically contain more keypoints.}
\label{fig:oe_keypoint_analysis}
\end{figure}

\section{DisastQA: Disaster Management Question Answer Benchmark}
\subsection{Overview}
\label{sec:overview}
We introduce \textbf{DisastQA}, a large-scale benchmark for disaster-response QA comprising two complementary tasks: DisastQA-MCQ for multiple-choice and DisastQA-OE for open-ended questions.
These tasks collectively span eight diverse real-world disaster event types with balanced coverage across categories (details in Appendix~\ref{app:dataset_details}).

MCQ items evaluate discriminative reasoning by requiring models to identify the correct answer among plausible, human-authored distractors, whereas OE items assess factual completeness through free-form responses reflecting real-world information needs in disaster management.

Construction follows a Human–LLM Collaboration Pipeline (Figure~\ref{fig:pipeline}) and is formally summarized in Algorithm~\ref{alg:qa_construction}. 
This hybrid approach strategically balances scalability with the factual rigor required for a safety-critical domain, consistent with prior visions of collective human--machine intelligence~\citep{li2025agentic}.
While LLMs provide scalable initial drafts, rigorous human expert oversight remains indispensable for rewriting, verification, and ensuring factual trustworthiness (see Appendix~\ref{app:human_refinement} for detailed error analysis and refinement statistics). This design, detailed in the following sections, ensures \textsc{DisastQA} is both large-scale and reliable.

\subsection{Annotator Expertise}
\label{sec:annotators}

All annotation and verification were performed by a team of five researchers with backgrounds in disaster resilience, crisis informatics, and infrastructure systems. The annotators possessed deep expertise across multiple disaster domains, ensuring accurate interpretation of diverse event types. Such domain-specific knowledge is essential for both verifying the factual accuracy of technical content and authoring plausible, context-aware distractors. They oversaw the entire quality-control process as well as the final validation of all items. Detailed annotation procedures for MCQ and OE tasks are described in Sections~\ref{sec:mcq_construction} and~\ref{sec:oe_construction}.

\subsection{DisastQA-MCQ Construction}
\label{sec:mcq_construction}

DisastQA-MCQ is constructed through a three-stage Human--LLM collaboration framework,
employing an \textbf{evidence-grounded generation strategy} with strict factual
constraints to convert disaster-related search queries into high-quality
multiple-choice questions.
We leverage the corpus from DisastIR~\citep{yin2025disastir}, a domain-specific
benchmark for crisis information seeking, primarily as a source of authoritative
evidence passages, ensuring that all questions are rooted in real-world disaster
reports rather than synthesized abstractions.

\subsubsection{Query-conditioned Question Generation}
\label{sec:query_rewrite}

Since the raw source queries are often short and keyword-based (e.g., ``earthquake bridge collapse'', ``flood evacuation route''), they are unsuitable for direct use in QA.
To address this, we prompt an LLM to rewrite each query into a well-formed, information-seeking question, leveraging synthetic generation techniques for domain adaptation~\citep{ma2021zero}.
These rewritten questions are conditioned on passages annotated as highly relevant (i.e., a score of 3 on a 0--3 scale).
Selecting only these top-tier passages ensures that rewritten questions are structurally grounded in definitive evidence rather than tangential mentions.
This step guarantees balanced coverage across the eight disaster categories and establishes a reliable factual basis for subsequent answer generation.

\subsubsection{Answer Option Generation} \label{sec:candidate_generation}
For each rewritten question--passage pair, the LLM generates one correct answer directly supported by the passage and three distractors that are semantically plausible but factually incorrect. These distractors are initially produced via strategic alteration of critical factual attributes (e.g., event date, affected location, magnitude, or policy detail). This stage captures the model's ability to generate diverse and competitive options, forming the preliminary multiple-choice structure. However, our pilot analysis revealed that LLM-generated distractors, while scalable, often lack sufficient discriminative power or remain trivially ``easy'' (e.g., simple negations, contextually irrelevant options). This finding underscores the necessity of the subsequent human refinement stage (Section~\ref{sec:human_refinement}) to systematically re-author and verify these options, ensuring all distractors are both challenging and realistic.

\subsubsection{Human Expert Refinement}
\label{sec:human_refinement}

Human experts then refine all question--answer sets to ensure factual accuracy, clarity, and difficulty balance.
Annotators verify that the gold answer is strictly supported by the passage and \textbf{manually revise distractors to ensure they are semantically plausible yet factually incorrect} (see Table~\ref{tab:error_refinement} in Appendix~\ref{app:error_refine_compare} for examples).
Each MCQ is independently validated by two annotators to confirm that exactly one option is correct; any disagreement is resolved by consensus.
Additionally, correct answers are uniformly distributed across the four option slots (A/B/C/D) to eliminate positional bias.

This rigorous refinement ensures that DisastQA-MCQ evaluates factual discrimination rather than mere surface-level memorization.
Quality control analysis reveals a human verification consistency of 95.4\% (954/1000 sampled items), where human experts independently solved the questions and matched the generated gold answers.
This high consistency confirms that the distractors are unambiguous and the gold answers are objectively correct.
Detailed prompt templates, filtering statistics, and annotation guidelines are provided in Appendix~\ref{appendix:guidelines} and~\ref{appendix:filter_stats}.

\subsection{DisastQA-OE Construction}
\label{sec:oe_construction}

\subsubsection{Query-conditioned Question Generation}
Similar to the MCQ track, the construction of OE questions begins with a DisastIR query--passage pair $(q, p_g)$, where $p_g$ is the most relevant passage (relevance score = 3). The LLM first rewrites the query into a natural-language question grounded in $p_g$. Human annotators then rigorously refine this draft for clarity, answerability, and factual fidelity. Unlike the MCQ track, this stage omits option generation and focuses solely on producing a concise, unambiguous question that elicits an explanatory, evidence-grounded response. Prompt templates for this stage are listed in Appendix~\ref{appendix:prompt}.

\begin{table}[t]
\centering
\small
\setlength{\tabcolsep}{3.5pt} 
\renewcommand{\arraystretch}{1.05}
\resizebox{\columnwidth}{!}{
\begin{tabular}{lcc}
\toprule
\textbf{Statistic} & \textbf{MCQ} & \textbf{OE} \\
\midrule
\# Instances           & 2,000 & 1,000 \\
Question Length        & 15.1±3.3 [5,32]  & 31.1±10.4 [7,78] \\
Option Length          & 9.5±3.3 [2,26]   & -- \\
Answer Length          & --    & 103.4±52.5 [11,350] \\
Keypoint Count         & --    & 4.4±1.5 [1,14] \\
\bottomrule
\end{tabular}}
\caption{
Descriptive statistics for the DisastQA-MCQ and DisastQA-OE tasks. 
All lengths are reported in tokens as mean±std [min,max].}
\label{tab:dataset_stats}
\end{table}

\subsubsection{Answer Construction}
For each verified question, annotators composed a gold reference answer ($a_{ref}$) grounded in the passage. 
While LLMs were optionally used to generate preliminary drafts, all reference answers were rewritten and verified by human experts to ensure high factual accuracy, conciseness, and neutrality. This approach maintains pipeline consistency with the MCQ track while also eliminating residual model bias or any unsupported content, yielding reliable and verifiable gold references, which in turn serve as the foundation for subsequent keypoint decomposition and fact-level evaluation.

\subsubsection{Keypoint Annotation}
To enable fine-grained factual evaluation, a 200-item subset of reference answers was decomposed into minimal factual units called \textbf{keypoints} ($\mathcal{K}$). For instance, a reference answer describing emergency safety actions is decomposed into atomic units such as ``(1) issue interim safety guidance'' and ``(2) coordinate with local fire services,'' as illustrated in Table~\ref{tab:keypoint_examples} (Appendix).
Since these keypoints serve as the ground truth for our evaluation metric, we employed a fully manual annotation process to ensure maximum reliability and eliminate LLM-induced biases or inconsistencies. 

To ensure reliability without relying on single-annotator subjectivity, we adopted a \textbf{strict double-pass protocol with consensus resolution}. Specifically, all reference answers were first annotated by primary experts and then reviewed by a verifier. Any discrepancies in keypoint boundaries were resolved through discussion to establish the final ground truth.
Statistical analysis of the final dataset shows a consistent granularity with an average of 4.39 keypoints per answer (SD=1.55), confirming stable annotation standards across disaster types (see Appendix~\ref{appendix:protocol} for detailed annotation guidelines). This annotation infrastructure underpins our primary metric, \textbf{Keypoint Coverage} (formally defined in Section~\ref{sec:oe_eval}), which evaluates models based on their recall of these essential atomic facts rather than surface-level lexical overlap.

\section{Dataset Statistics and Analysis}
\label{sec:dataset_stats}
We analyze the statistical properties of DisastQA, focusing on the impact of human refinement, descriptive trends, and domain complexity.

\subsection{Descriptive Statistics}
The final \textsc{DisastQA} dataset comprises 3,000 instances, including 2,000 multiple-choice and 1,000 open-ended questions, with balanced coverage across eight disaster types (see Table~\ref{tab:dataset_stats}).
MCQ items are precision-oriented (avg.\ 15.1 tokens), assessing discriminative reasoning over specific facts.
In contrast, OE items feature longer questions (avg.\ 31.1 tokens) that require comprehensive, human-authored reference answers (avg.\ 103.4 tokens).
Within the annotated subset, each reference answer is decomposed into an average of 4.4 atomic keypoints, serving as a quantifiable measure of factual complexity and granular information density.

\subsection{Domain-Specific Complexity Analysis}
The keypoint annotations quantitatively reveal the high factual density that characterizes disaster-response QA.
As shown in Figure~\ref{fig:oe_keypoint_analysis}, while many questions require reasoning over 3--5 facts, a significant portion necessitate integrating 8 or more atomic facts, often dispersed throughout the context, representing a level of complexity not typically captured in general-domain QA benchmarks.

This high factual density underscores the distinct challenges of the disaster domain. This validates the necessity of our keypoint-based evaluation protocol, as traditional metrics such as ROUGE often fail to penalize models that produce superficially fluent yet factually incomplete answers, a limitation widely recognized in recent generation benchmarks~\citep{scialom2021questeval}.

\section{Evaluation Methodology}
\label{sec:evaluation_methodology}

We denote $a^\star$ as the gold answer for MCQ, $\mathcal{K}$ as the gold keypoints for OE, and $a_{gen}$ as the model-generated answer.
To ensure systematic and interpretable evaluation, we adopt a unified methodology disentangling a model’s parametric knowledge from its ability to utilize external evidence.
Our core innovation is a three-context evaluation framework comprising \textit{Base}, \textit{Mix}, and \textit{Golden}, which builds upon prior work on robustQA~\citep{han-etal-2023-robustqa}.
This framework allows us to assess model behavior not only under idealized conditions but also under realistic uncertainty, reflecting how disaster information is often incomplete, noisy, or misleading.
We compare these results with general-domain benchmarks such as Natural Questions~\citep{Kwiatkowski2019natural} (Section~\ref{sec:results}) to contextualize disaster-domain robustness.

\subsection{Evaluation Settings}
\label{sec:eval_settings}
To disentangle parametric knowledge from reasoning capabilities, we evaluate models under three controlled information settings (see qualitative examples in Appendix~\ref{appendix:prompt}):

\textbf{Base (Closed-Book).}
The model answers the question $q$ using only its internal parametric knowledge, serving as a baseline for hallucination and internal knowledge reliability.

\textbf{Golden (Oracle).}
The model receives the question $q$ and ground-truth passage $p_g$, representing an oracle condition with perfect evidence and measuring upper-bound reasoning performance.

\textbf{Mix (Noisy Context).}
The context $p_c$ combines the golden passage $p_g$ with $k=4$ stratified distractors ($P_{dist}$), simulating a standard \textbf{Top-5 retrieval} scenario ($1$ Gold + $4$ Noise)~\citep{karpukhin2020dense, chen2024benchmarking}.
Rather than evaluating retrieval performance, we quantify the LLM’s discriminative ability to identify and utilize relevant evidence under high-ranking noise.
Passages are presented in randomized order to prevent position bias.

\subsection{MCQ Evaluation}
\label{sec:mcq_eval}

Each MCQ instance consists of a question $x$, four options 
$\{o_A,o_B,o_C,o_D\}$ with one correct answer $a^\star$, 
and a passage $p_c$ under the chosen evaluation context. 
Models are required to output the option.

\paragraph{Accuracy Metric.}
Evaluation is based on \textit{exact-match accuracy}, i.e., the proportion of test questions for which the predicted option matches the gold answer. Since each item is validated to ensure exactly one correct answer, accuracy directly reflects a model’s ability to discriminate correctness.

\subsection{OE Evaluation}
\label{sec:oe_eval}
Each OE instance consists of a question $x$, a passage $p_c$, and a human reference answer $a_{ref}$.
To rigorously evaluate factual adequacy, we adopt a \textbf{Human-Verified Keypoint Protocol}.

\paragraph{Keypoint Decomposition.}
Domain experts decompose each reference answer $a_{ref}$ into a set of atomic keypoints
$\mathcal{K} = \{k_1, k_2, \dots, k_n\}$, aligning with the atomic fact definition in FActScore~\citep{min2023factscore}, where each keypoint represents a minimal factual unit required for correctness.

\paragraph{Scoring Procedure (Human Verification).}
We conduct a fully human-annotated evaluation on a representative 200-item OE subset.
To ensure structural balance and representativeness, we employ \textbf{Stratified Random Sampling}: we randomly sample exactly 25 questions from each of the 8 disaster types (Total: $25 \times 8 = 200$).
All 20 models are evaluated on this identical subset to guarantee fair, paired comparisons. Annotators are presented with the model-generated answer $a_{gen}$ and the gold keypoints $\mathcal{K}$.
For each keypoint $k_i \in \mathcal{K}$, the annotator determines whether its semantic meaning
is entailed by $a_{gen}$, regardless of surface phrasing.
This binary judgment $\mathbb{I}(k_i \in a_{gen})$ equals 1 if the fact is present and 0 otherwise.
Detailed human annotation guidelines are provided in Appendix~\ref{appendix:guidelines}.

We define Keypoint Coverage as a metric of strict factual recall:
\begin{equation}
\label{eq:coverage}
\text{Coverage}(a_{gen}) =
\frac{1}{|\mathcal{K}|} \sum_{k \in \mathcal{K}} \mathbb{I}(k \in a_{gen}).
\end{equation}

This human-verified evaluation prioritizes factual completeness and avoids rewarding verbosity
or surface-level lexical overlap, making it particularly suitable for risk-sensitive
disaster QA settings.

\begin{table*}[t]

\centering
\small
\setlength{\tabcolsep}{4pt}
\renewcommand{\arraystretch}{1.05}

\resizebox{\textwidth}{!}{%
\begin{tabular}{ll|ccc|ccc|ccc|ccc|ccc}
\toprule
\multirow{4}{*}{\textbf{Model}} & 
\multirow{4}{*}{\textbf{Params}} & 
\multicolumn{3}{c|}{\textbf{MCQ}} & 
\multicolumn{12}{c}{\textbf{OE}} \\
\cmidrule(lr){3-5} \cmidrule(lr){6-17}
 &  & \multicolumn{3}{c|}{\textbf{Accuracy}} & 
\multicolumn{3}{c|}{\textbf{ROUGE-L}} & 
\multicolumn{3}{c|}{\textbf{BLEU-4}} & 
\multicolumn{3}{c|}{\textbf{BERTScore-F1}} & 
\multicolumn{3}{c}{\textbf{Coverage (\%)}} \\
\cmidrule(lr){3-5} \cmidrule(lr){6-8} \cmidrule(lr){9-11} \cmidrule(lr){12-14} \cmidrule(lr){15-17}
 &  & Base & Mix & Golden & Base & Mix & Golden & Base & Mix & Golden & Base & Mix & Golden & Base & Mix & Golden \\
\midrule
GPT-5.2              & --       & \textbf{0.9314} & \textbf{0.9670} & \textbf{0.9965} & 0.178 & 0.294 & 0.286 & 0.026 & 0.074 & 0.063 & 0.854 & 0.886 & 0.882 & \textbf{87.1} & \textbf{93.5} & 94.6 \\
Gemini-3 Pro         & --       & 0.9219 & 0.9535 & 0.9670 & 0.210 & \textbf{0.426} & 0.437 & 0.043 & 0.196 & 0.193 & 0.868 & \textbf{0.913} & 0.919 & 84.7 & 91.5 & \textbf{96.5} \\
GPT-4o               & --       & 0.9105 & 0.9625 & 0.9935 & 0.229 & 0.381 & 0.367 & 0.069 & 0.174 & 0.160 & 0.878 & 0.910 & 0.910 & 83.7 & 89.8 & 95.4 \\
Gemini-1.5 Pro       & --       & 0.8895 & 0.9380 & 0.9870 & 0.205 & 0.399 & 0.317 & 0.038 & 0.178 & 0.094 & 0.868 & 0.906 & 0.897 & 84.5 & 84.0 & 95.1 \\
Qwen-3-8B            & 8.19B    & 0.8865 & 0.9625 & \textbf{0.9965} & 0.234 & 0.332 & 0.357 & 0.063 & 0.129 & 0.139 & 0.878 & 0.896 & 0.906 & 79.1 & 87.1 & 94.0 \\
Yi-6B-Chat           & 6.06B    & 0.8575 & 0.9565 & 0.9930 & 0.257 & 0.305 & 0.358 & 0.090 & 0.126 & 0.167 & 0.879 & 0.892 & 0.900 & 73.0 & 84.7 & 86.5 \\
Llama-3-8B           & 8.03B    & 0.8715 & 0.9565 & 0.9910 & \textbf{0.273} & 0.378 & 0.425 & \textbf{0.104} & \textbf{0.202} & 0.223 & \textbf{0.885} & 0.902 & 0.917 & 77.5 & 85.0 & 93.4 \\
Qwen-3-4B            & 4.02B    & 0.8605 & 0.9600 & 0.9960 & 0.234 & 0.337 & 0.356 & 0.057 & 0.125 & 0.127 & 0.879 & 0.897 & 0.907 & 76.5 & 83.5 & 91.7 \\
Qwen-2.5-3B-Instr.   & 3.00B    & 0.8530 & 0.9545 & 0.9915 & 0.236 & 0.339 & 0.364 & 0.061 & 0.129 & 0.134 & 0.880 & 0.899 & 0.906 & 76.9 & 85.4 & 93.4 \\
Phi-2                & 2.78B    & 0.8725 & 0.9570 & 0.9895 & 0.251 & 0.323 & 0.355 & 0.088 & 0.154 & 0.169 & 0.882 & 0.897 & 0.905 & 74.0 & 82.0 & 86.9 \\
Mistral-7B-Instr.    & 7.24B    & 0.8180 & 0.9435 & 0.9865 & 0.243 & 0.328 & 0.384 & 0.072 & 0.145 & 0.168 & 0.879 & 0.896 & 0.911 & 76.3 & 85.4 & 91.6 \\
Gemma-7B             & 8.54B    & 0.8025 & 0.8975 & 0.9805 & 0.263 & 0.386 & \textbf{0.467} & 0.088 & 0.189 & \textbf{0.242} & 0.879 & 0.907 & \textbf{0.922} & 72.9 & 69.7 & 89.7 \\
DeepSeek-v3-7B       & 7.00B    & 0.8020 & 0.9435 & 0.9860 & 0.270 & 0.377 & 0.422 & 0.100 & 0.193 & 0.221 & 0.884 & 0.906 & 0.915 & 73.5 & 84.1 & 92.0 \\
Llama-3.2-3B-Instr.  & 3.21B    & 0.8510 & 0.9550 & 0.9865 & \textbf{0.273} & 0.374 & 0.418 & 0.094 & 0.177 & 0.201 & 0.884 & 0.905 & 0.916 & 76.6 & 84.3 & 92.0 \\
Llama-3.2-1B-Instr.  & 1.24B    & 0.8120 & 0.9345 & 0.9690 & 0.256 & 0.326 & 0.376 & 0.091 & 0.156 & 0.190 & 0.881 & 0.897 & 0.909 & 74.5 & 86.5 & 90.4 \\
Falcon-3-1B-Instr.   & 1.67B    & 0.7475 & 0.9120 & 0.9690 & 0.248 & 0.343 & 0.401 & 0.078 & 0.157 & 0.174 & 0.882 & 0.900 & 0.914 & 71.1 & 84.1 & 92.3 \\
AceMath-1.5B-Instr.  & 1.78B    & 0.7410 & 0.9210 & 0.9815 & 0.244 & 0.342 & 0.373 & 0.082 & 0.167 & 0.179 & 0.873 & 0.899 & 0.897 & 67.9 & 86.6 & 92.3 \\
Hunyuan-4B-Instr.    & 4.22B    & 0.8195 & 0.9435 & 0.9780 & 0.192 & 0.278 & 0.284 & 0.043 & 0.106 & 0.100 & 0.861 & 0.885 & 0.884 & 67.2 & 84.5 & 91.4 \\
Hunyuan-7B-Instr.    & 7.50B    & 0.5700 & 0.8790 & 0.9305 & 0.123 & 0.323 & 0.266 & 0.012 & 0.127 & 0.083 & 0.816 & 0.892 & 0.881 & 39.6 & 72.6 & 79.9 \\
Qwen-3-0.6B          & 0.75B    & 0.7455 & 0.8290 & 0.8810 & 0.242 & 0.289 & 0.314 & 0.077 & 0.129 & 0.123 & 0.869 & 0.870 & 0.875 & 69.7 & 85.7 & 90.1 \\
\bottomrule

\end{tabular}}
\caption{
Summary of model performance on \textsc{DisastQA}-MCQ and OE tasks across three retrieval settings (Base, Mix, Golden).
Metrics include Accuracy for MCQ, and ROUGE-L, BLEU-4, BERTScore-F1, and Keypoint Coverage for OE evaluation.
Best results in each column are highlighted in \textbf{bold}.
}
\label{tab:mcq_oe_results}
\end{table*}
\paragraph{Complementary Metrics.}
For comparability with prior work, we report ROUGE-L,
BLEU-4, and BERTScore-F1 on all 1,000 OE questions.
These automated metrics are secondary indicators of surface-level similarity,
while human-verified Keypoint Coverage serves as the primary measure of factual adequacy.

\begin{figure}[t]
\centering
\includegraphics[width=\columnwidth]{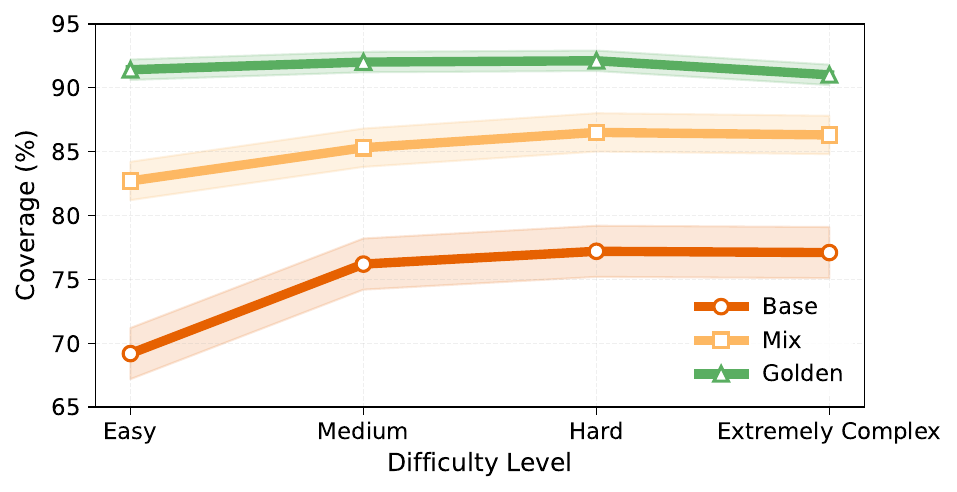}
\caption{
Keypoint Coverage across difficulty levels under Base, Mix, and Golden.
}
\label{fig:oe_difficulty_curve}
\end{figure}

\subsection{Summary of Experimental Setup}
In summary, DisastQA integrates \textbf{Accuracy} for MCQ and \textbf{Keypoint Coverage} for OE.  
Together with controlled retrieval settings, these metrics ensure that our benchmark captures (1) discriminative reasoning, (2) factual adequacy, and (3) evidence utilization under uncertainty. 
To ensure reproducibility, we adopt a deterministic decoding strategy for all generative tasks, eliminating randomness to enable fair comparison.

\subsection{Evaluated Models}
We evaluate 20 models, spanning diverse scales from efficient open-weight models (0.6B--8B) to proprietary frontier systems. We strictly prioritize instruction-tuned variants to align with conversational deployment needs.
Our selection covers major open families (\textit{Qwen, Llama, Gemma, Mistral, Phi, Falcon, Yi, DeepSeek, Hunyuan}) and proprietary APIs, including the latest state-of-the-art versions (\textit{GPT-5.2}, \textit{Gemini-3 Pro}). 
This selection strategy allows us to benchmark the \textasciitilde7B parameter class, a standard for local resource-constrained disaster scenarios, against the performance upper bounds established by proprietary SOTAs, revealing the critical trade-off between practical deployment efficiency and robust reasoning reliability.
Detailed implementation notes and model sources are provided in Appendix~\ref{appendix:model_details}.

\section{Results and Analysis}
\label{sec:results}

\subsection{Overall Performance on DisastQA-MCQ}
Table~\ref{tab:mcq_oe_results} summarizes the performance of 20 models. Across all settings, we observe a consistent hierarchy of \textit{Base} < \textit{Mix} < \textit{Golden}, confirming that \textsc{DisastQA} questions require external evidence, underscoring the \textbf{critical role of context retrieval}.

\paragraph{Gap between Open-Source and Proprietary Models.}
In the \textit{Base} setting, the latest frontier model GPT-5.2 establishes a new upper bound with 93.1\% accuracy.
Notably, the open-weight model Qwen-3-8B performs exceptionally well, narrowing the gap with proprietary systems in the \textit{Golden} setting.
However, in the \textit{Base} (no-context) scenario, \textbf{GPT-5.2 retains a substantial advantage} (93.1\% vs.\ Qwen-3’s 88.7\%), indicating that proprietary models still possess stronger internalized world knowledge derived from massive pre-training.

\paragraph{Robustness to Contextual Noise.}
Comparing \textit{Golden} and \textit{Mix} reveals how models handle noisy evidence.
While GPT-5.2 achieves the highest absolute accuracy in \textit{Mix}, \textbf{Gemini-3 Pro exhibits the strongest relative robustness}, showing the smallest drop from \textit{Golden} to \textit{Mix}.
This indicates that although GPT-5.2 excels in precision, Gemini-3 Pro is particularly stable in filtering irrelevant or conflicting passages, an essential capability for real-world decision-support scenarios.

\subsection{Overall Performance on DisastQA-OE}

\paragraph{Trade-off: Fluency vs.\ Factuality.}
Results reveal a clear \textbf{trade-off between fluency and factual completeness}.
As shown in Table~\ref{tab:mcq_oe_results}, Gemma-7B achieves high ROUGE-L scores, reflecting text fluency, yet its Keypoint Coverage lags behind.
By contrast, Gemini-3 Pro demonstrates superior factual stability.
While smaller models often struggle to maintain information density in longer responses, Gemini-3 Pro consistently synthesizes comprehensive, multi-fact answers.
This finding is critical: whereas GPT-5.2 dominates discriminative tasks (MCQ), Gemini-3 Pro exhibits the generative capabilities required for high-fidelity disaster reporting.

\subsection{Performance by Complexity}
Figure~\ref{fig:oe_difficulty_curve} illustrates the aggregate performance trends across difficulty levels, while detailed per-model numerical results are provided in Table~\ref{tab:coverage_difficulty_models_all} (Appendix~\ref{appendix:difficulty_results}).
The dataset distribution is skewed (see Table~\ref{tab:difficulty_stats}), with ``Extremely Complex'' items representing a long-tail subset ($N=26$).
Due to this scarcity, results on this tier should be interpreted as exploratory trends rather than statistically conclusive benchmarks.
However, the relative degradation trend offers a robust signal:
While frontier models like GPT-5.2 and Gemini-3 Pro maintain high performance on complex queries, \textbf{smaller models degrade sharply}.
For instance, Qwen-3-8B drops from 94.2\% on Easy items to 87.5\% on Extremely Complex ones in the \textit{Golden} setting.
This suggests that while smaller models are competent on simple facts, they lack the reasoning depth to integrate multiple evidence sources to derive correct conclusions.
This fragility is exacerbated in the \textit{Mix} setting, where \textbf{reasoning complexity amplifies the distraction effect}, causing a sharp performance degradation for sub-10B models.

\subsection{Comparison with General-Domain Benchmarks}
To test domain transfer, we compare \textsc{DisastQA}-MCQ against an 8-subject subset of MMLU-Pro~\cite{wang2024mmlu-pro} (details in Appendix~\ref{appendix:mmlu_results}).
Figure~\ref{fig:mcq_performance} illustrates the ranking divergence.
While GPT-5.2 leads on both, models like DeepSeek-v3-7B drop significantly in disaster contexts compared to general benchmarks.
This confirms that \textbf{general-domain performance does not guarantee reliability} in safety-critical scenarios.
\begin{table}[t]
\centering
\small
\setlength{\tabcolsep}{5pt}
\renewcommand{\arraystretch}{1.05}
\begin{tabular}{lrr}
\toprule
\textbf{Difficulty} & \textbf{\#Q} & \textbf{Proportion} \\
\midrule
Easy & 109 & 10.9\% \\
Medium & 684 & 68.4\% \\
Hard & 181 & 18.1\% \\
Extremely Complex & 26 & 2.6\% \\
\midrule
Total & 1{,}000 & 100\% \\
\bottomrule
\end{tabular}
\caption{
\textbf{Distribution of difficulty levels for open-ended questions.} 
Tiers are defined by keypoint (KP) counts: Easy (1--2 KP), Medium (3--5 KP), Hard (6--7 KP), and Extremely Complex (8+ KP).}
\label{tab:difficulty_stats}
\end{table}

\subsection{Event Type Breakdown}
\label{sec:event_breakdown}
As illustrated in Appendix~\ref{appendix:event_breakdown} (Figure~\ref{fig:event_breakdown_appendix}), there is substantial performance variance across disaster types (Table~\ref{tab:mcq_event_all}). In the \textit{Golden} setting, GPT-5.2 achieves near-saturated performance ($>$99\%) across all categories.
However, domain gaps remain for open-weight models.
For instance, in the \textit{Base} setting, DeepSeek-v3-7B drops to 78.8\% accuracy on ``Biological'' vs. 82.0\% on ``Geohazards''.
\textbf{Technological and Biological disasters are consistently the most challenging} due to specialized terminology (e.g., chemical compounds, pathogen details) that is underrepresented in pre-training data.

\subsection{Error Analysis}
We conducted a qualitative analysis of failure cases (see Appendix~\ref{app:human_refinement}).
For MCQs, the dominant failure mode is distractor attraction, where models are misled by incorrect options sharing high lexical overlap with the question. This reveals reliance on \textbf{spurious keyword matching rather than multi-hop verification}.
For OEs, even GPT-5.2 occasionally exhibits partial coverage, omitting quantitative details (e.g., casualty figures) while capturing the general narrative.
This underscores the value of Keypoint Coverage as an evaluation signal for identifying
\textbf{factual omissions not captured by surface-level metrics}.

\section{Conclusion}
We introduce \textbf{DisastQA}, a large-scale, human-verified benchmark designed to evaluate model reliability under disaster-response constraints.
By systematically varying evidence quality, we uncover a critical robustness gap: models excelling with perfect contexts often fail to reject noise in realistic scenarios.
Furthermore, we find that surface metrics such as ROUGE overestimate factual correctness compared to keypoint-based evaluation, masking significant reliability failures.
Ultimately, our results show that strong general-domain capabilities do not guarantee reliability in safety-critical settings.
DisastQA thus serves as a critical stress test, shifting the paradigm from evaluating conversational fluency to ensuring verifiable, evidence-grounded robustness in high-stakes environments.

\section*{Limitations}
\label{sec:limitations}

DisastQA currently focuses on \textbf{text-based QA in English}, prioritizing the rigorous evaluation of factual precision and evidence grounding.
This scope allows us to isolate and quantify core reasoning capabilities without the confounding factors of additional modalities.
While real-world disaster management involves diverse information streams, extending the framework to include \textit{multimodal} or \textit{multilingual} elements represents a natural direction for future research to further broaden the benchmark's applicability.

\section*{Ethics Statement}
\label{sec:ethics}

The \textsc{DisastQA} dataset is constructed from publicly available
disaster-related reports intended for information sharing.
All released data consist only of derived annotations (questions, answers,
and keypoints), with source references provided where applicable.
Automated and manual screening was applied to remove any Personally
Identifiable Information (PII).

Human annotators involved in the dataset refinement were compensated fairly
and informed of the nature of the task.
The dataset is released solely for non-commercial research purposes.
Models evaluated in this work are intended for research benchmarking
and should not be used as autonomous systems in safety-critical
decision-making without appropriate human oversight.
In particular, erroneous or incomplete answers produced by QA systems
in disaster contexts may lead to misinformed decisions, highlighting the
importance of treating benchmark results as diagnostic signals rather than
deployment-ready guarantees.

\section*{Acknowledgements}
This work used the ACES at Texas A\&M University, DeltaAI, and Delta GPU resources at the National Center for Supercomputing Applications through allocations CIV250019 and CIV250021 from the Advanced Cyberinfrastructure Coordination Ecosystem: Services \& Support (ACCESS) program, which is supported by U.S. National Science Foundation grants \#2138259, \#2138286, \#2138307, \#2137603, and \#2138296.
We also acknowledge the use of high-performance computing resources provided by the Texas A\&M University High Performance Research Computing (HPRC) facility.

\bibliography{custom}

\clearpage
\appendix
\section{Dataset Details}
\label{app:dataset_details}

\paragraph{Overview.}
DisastQA covers eight major disaster event types:
\textit{meteorological, geophysical, hydrological, climatological, biological, technological, extraterrestrial, and conflict-induced}.  
Each event type includes both Multiple-Choice (MCQ) and Open-Ended (OE) questions, 
ensuring balanced coverage across task formats and categories.  
Table~\ref{tab:dataset_details} below summarizes the distribution across all event types.

\begin{table}[h]
\centering
\small
\begin{tabular}{lcc}
\toprule
\textbf{Event Type} & \textbf{MCQ Count} & \textbf{OE Count} \\
\midrule
Meteorological      & 250 & 125 \\
Geophysical         & 250 & 125 \\
Hydrological        & 250 & 125 \\
Climatological      & 250 & 125 \\
Biological          & 250 & 125 \\
Technological       & 250 & 125 \\
Extraterrestrial    & 250 & 125 \\
Conflict-induced    & 250 & 125 \\
\midrule
\textbf{Total}      & 2000 & 1000 \\
\bottomrule
\end{tabular}
\caption{
Distribution of DisastQA questions across disaster event types.  
Each event type contains an equal number of questions to maintain balance across MCQ and OE tasks.
}
\label{tab:dataset_details}
\end{table}

\section{Human Refinement and Quality Control}
\label{app:human_refinement}

This section details the human refinement process that ensured factual precision, plausibility, and clarity across all DisastQA items.  
Annotators systematically rewrote and verified a substantial portion of LLM-generated drafts to ensure correctness, grounding, and answerability.  
Each MCQ underwent independent solve-check validation by two annotators, while OE items were double-reviewed through a keypoint consensus protocol to guarantee factual completeness and consistency.

\subsection{The Necessity of Human Refinement: An Error Analysis}
\label{sec:human_refinement_analysis}

Although LLMs can generate scalable drafts, we observe recurring failure modes that highlight the necessity of human refinement. The most common issues include distractors lacking discriminative power (approx. 25\% of MCQ drafts), factual inaccuracies (15–20\%), ambiguous wording (10–15\%), and incomplete factual coverage in open-ended answers (about 20–25\%). These error patterns underscore that while LLMs accelerate data generation, their raw outputs fall short of the reliability standards required for safety-critical domains. Human annotators systematically corrected these issues, ensuring that all benchmark items are precise, passage-grounded, and factually complete. \textit{Representative failure cases and a detailed breakdown of LLM failure types are provided in Appendix~\ref{appendix:error_cases}.}

\subsection{Refinement Statistics}
\label{app:refinement_stats}

Table~\ref{tab:rewrite_stats} summarizes the quantitative extent of human intervention across dataset components.  
The \textit{rewrite rate} denotes the proportion of LLM drafts that were modified by human annotators due to ambiguity, weak grounding, or factual inaccuracies.  
For Open-Ended (OE) tasks, while LLMs produced preliminary drafts, all gold reference answers were subsequently rewritten and verified by human experts to ensure factual accuracy, conciseness, and passage fidelity, thereby eliminating residual model bias in the final dataset.

\begin{table}[h]
\centering
\small
\setlength{\tabcolsep}{3pt}
\renewcommand{\arraystretch}{1.05}
\resizebox{\columnwidth}{!}{%
\begin{tabular}{lcc}
\toprule
\textbf{Task Component} & \textbf{Rewrite Rate} & \textbf{QC Protocol} \\
\midrule
MCQ Questions      & 35\%  & Clarity + Solve-check \\
MCQ Distractors    & 45\%  & Plausibility validation \\
OE Questions       & 30\%  & Clarity rewrite \\
OE Gold Answers    & 80–100\% & Human rewrite + factual verification \\
OE Keypoints       & --    & Double-pass consensus \\
\bottomrule
\end{tabular}}
\caption{
Human refinement statistics across dataset components.  
Rewrite rate denotes the fraction of LLM drafts revised during human quality control.
}
\label{tab:rewrite_stats}
\end{table}

\subsection{Draft–Refinement Examples}
\label{app:rewrite_examples}

Table~\ref{tab:rewrite_examples} presents representative examples of MCQ and OE refinements.  
Human intervention consistently improved factual precision, plausibility of distractors, and completeness of reference answers.

\begin{table*}[t]
\centering
\small
\setlength{\tabcolsep}{4pt}
\renewcommand{\arraystretch}{1.05}
\resizebox{\textwidth}{!}{%
\begin{tabular}{p{0.14\textwidth} p{0.41\textwidth} p{0.41\textwidth}}
\toprule
\textbf{Task} & \textbf{LLM Draft} & \textbf{Human Verification / Authoring} \\
\midrule
MCQ &
Q: ``What is one important precaution to follow when handling the Proflow\textsuperscript{TM} Hepatitis A Test Device?'' \newline
Opts: A. Store the device in direct sunlight to maintain its efficacy.; B. Ensure proper ventilation and wear personal protective equipment during use.; C. Dispose of the device in regular trash without special handling precautions.; D. Only use the device outdoors to avoid contamination. &
Q: ``What is one important precaution to follow when handling the Proflow\textsuperscript{TM} Hepatitis A Test Device?'' \newline
Opts: A. Store the device in direct sunlight to maintain its efficacy.; B. \textbf{Ensure proper ventilation and wear personal protective equipment during use.}; C. Dispose of the device in regular trash without special handling precautions.; D. Only use the device outdoors to avoid contamination. \\
\midrule
MCQ &
Q: ``Which of the following practices should local governments implement to enhance their floodplain management under the National Flood Insurance Program?'' \newline
Opts: A. Establishing a program to subsidize insurance for all property owners regardless of flood risk.; B. Implementing a mandatory evacuation plan for all residents during flood warnings.; C. Revising the floodplain ordinance to include cumulative substantial damage requirements.; D. Providing tax breaks for property development in flood-prone areas. &
Q: ``Which of the following practices should local governments implement to enhance their floodplain management under the National Flood Insurance Program?'' \newline
Opts: A. Establishing a program to subsidize insurance for all property owners regardless of flood risk.; B. Implementing a mandatory evacuation plan for all residents during flood warnings.; C. \textbf{Revising the floodplain ordinance to include cumulative substantial damage requirements.}; D. Providing tax breaks for property development in flood-prone areas. \\
\midrule
OE &
\textit{(LLM question and answer draft)} \newline
Q: ``What things that people do cause acid rain to happen?'' \newline
A: ``The passage says acid rain is made because people burn fuels like coal and oil in factories and cars.'' &
Q: ``What are the primary human activities that lead to the formation of acid rain?'' \newline
\textbf{Gold Answer (human-authored):} ``The primary human activities leading to the formation of acid rain are the burning of fossil fuels in power plants and vehicles, as stated in the passage.'' \newline
\textbf{Keypoints (human-annotated):} \{Burning of fossil fuels in power plants.; Burning of fossil fuels in vehicles.\} \\
\midrule
OE &
\textit{(LLM question and answer draft)} \newline
Q: ``How does CBPP hurt the economy in places where it happens?'' \newline
A: ``CBPP makes cows sick and die, which causes big money loss for farmers and also hurts the country’s economy.'' &
Q: ``How does Contagious Bovine Pleuropneumonia (CBPP) affect the economic stability of households and national economies in regions where it is prevalent?'' \newline
\textbf{Gold Answer (human-authored):} ``CBPP induces significant economic losses and leads to serious livestock production problems, negatively influencing people’s livelihoods in affected countries.'' \newline
\textbf{Keypoints (human-annotated):} \{CBPP causes significant economic losses.; Serious livestock production problems.; Negatively impacts people’s livelihoods.\} \\
\bottomrule
\end{tabular}}
\caption{
Illustrative examples of MCQ and OE refinement under our Human-LLM pipeline. 
For \textbf{MCQ}, the LLM drafts question+options and humans refine/verify. 
For \textbf{OE}, the LLM drafts the question and answers; \textbf{the reference answers and keypoints are written/annotated by humans}.
}
\label{tab:rewrite_examples}
\end{table*}

This section provides a full breakdown of model performance across difficulty levels.
Following the keypoint-based scoring design in Section~\ref{sec:evaluation_methodology}, 
we categorize Open-Ended (OE) questions into four difficulty levels—\textbf{Easy}, \textbf{Medium}, \textbf{Hard}, and \textbf{Extremely Complex}—based on the number of keypoints per reference answer.
Table~\ref{tab:difficulty_full} reports Keypoint Coverage (\%) across the three evaluation settings 
(\textbf{Base}, \textbf{Mix}, and \textbf{Golden}) for all 20 evaluated models.

These results complement the main findings in Section~\ref{sec:results}, 
demonstrating how evidence quality and model capacity interact with task difficulty.
As shown, performance generally increases from Base → Mix → Golden for all difficulty levels, 
but degradation remains substantial when the number of keypoints (i.e., factual units) grows.

\begin{table*}[t]
\centering
\small
\setlength{\tabcolsep}{3pt}
\renewcommand{\arraystretch}{1.05}
\resizebox{\textwidth}{!}{%
\begin{tabular}{l|ccc|ccc|ccc|ccc}
\toprule
\multirow{2}{*}{\textbf{Model}} & 
\multicolumn{3}{c|}{\textbf{Easy}} & 
\multicolumn{3}{c|}{\textbf{Medium}} & 
\multicolumn{3}{c|}{\textbf{Hard}} & 
\multicolumn{3}{c}{\textbf{Extremely Complex}} \\
\cmidrule(lr){2-4} \cmidrule(lr){5-7} \cmidrule(lr){8-10} \cmidrule(lr){11-13}
 & Base & Mix & Golden & Base & Mix & Golden & Base & Mix & Golden & Base & Mix & Golden \\
\midrule
GPT-5.2              & \textbf{91.03} & 94.66 & 89.22 & \textbf{86.20} & 92.33 & 95.51 & \textbf{85.89} & \textbf{96.05} & 95.23 & \textbf{100.00} & 95.00 & \textbf{100.00} \\
Gemini-3 Pro         & 83.13 & 89.72 & 94.36 & 85.01 & \textbf{92.59} & \textbf{97.07} & 85.29 & 91.12 & \textbf{96.18} & 78.87 & 75.00 & \textbf{100.00} \\
GPT-4o               & 82.89 & \textbf{96.22} & 96.06 & 84.42 & 91.87 & 95.25 & 80.07 & 93.06 & 95.56 & 89.58 & 90.83 & \textbf{100.00} \\
Gemini-1.5 Pro       & 79.96 & 91.17 & 95.44 & 82.57 & 88.26 & 94.15 & 83.96 & 90.40 & 95.38 & 86.67 & 81.66 & 88.75 \\
Qwen-3-8B            & 76.35 & 87.26 & 94.17 & 78.37 & 89.64 & 94.46 & 76.50 & 93.80 & 93.62 & 65.00 & 74.17 & 87.50 \\
Yi-6B-Chat           & 67.48 & 85.71 & 86.50 & 71.72 & 84.87 & 86.23 & 76.85 & 83.04 & 88.65 & 95.83 & 96.43 & \textbf{100.00} \\
Llama-3-8B           & 74.99 & 84.52 & \textbf{96.28} & 75.31 & 88.75 & 91.90 & 82.21 & 90.09 & 91.02 & 85.42 & \textbf{100.00} & 83.03 \\
Qwen-3-4B            & 79.93 & 86.56 & 90.56 & 77.42 & 87.35 & 93.11 & 78.05 & 86.22 & 83.75 & 85.83 & 93.75 & 86.66 \\
Qwen-2.5-3B-Instr.   & 74.98 & 85.71 & 96.17 & 76.89 & 87.54 & 91.98 & 80.33 & 87.50 & 92.71 & 81.67 & 90.83 & 81.43 \\
Phi-2                & 75.80 & 81.10 & 87.27 & 69.63 & 81.87 & 89.78 & 76.46 & 83.46 & 86.67 & 68.75 & 80.18 & 88.75 \\
Mistral-7B-Instr.    & 76.00 & 85.34 & 88.88 & 75.67 & 84.59 & 89.99 & 78.82 & 81.28 & 88.27 & 82.50 & 77.08 & 90.83 \\
Gemma-7B             & 66.85 & 86.36 & 93.67 & 75.23 & 83.59 & 91.53 & 75.68 & 85.32 & 93.41 & 75.42 & 77.14 & 90.83 \\
DeepSeek-v3-7B       & 72.33 & 88.61 & 88.60 & 71.55 & 86.80 & 90.34 & 81.37 & 85.62 & 88.55 & 90.00 & 85.42 & \textbf{100.00} \\
Llama-3.2-3B-Instr.  & 66.18 & 86.25 & 89.37 & 76.41 & 85.12 & 94.04 & 77.02 & 87.62 & 88.89 & 88.75 & 95.83 & 95.00 \\
Llama-3.2-1B-Instr.  & 82.10 & 77.28 & 93.66 & 72.27 & 82.01 & 90.00 & 72.47 & 87.16 & 89.63 & 75.59 & 87.50 & 95.00 \\
Falcon-3-1B-Instr.   & 77.32 & 87.33 & 91.31 & 71.90 & 84.03 & 92.61 & 74.71 & 85.29 & 91.69 & 59.58 & 83.75 & 92.86 \\
AceMath-1.5B-Instr.  & 71.76 & 85.17 & 85.99 & 68.44 & 85.36 & 92.36 & 63.87 & 88.52 & 94.13 & 80.62 & 83.75 & 90.18 \\
Hunyuan-4B-Instr.    & 58.79 & 85.47 & 92.50 & 66.49 & 86.13 & 89.64 & 71.81 & 90.33 & 92.63 & 95.83 & 86.66 & 81.61 \\
Hunyuan-7B-Instruct  & 36.75 & 84.74 & 76.39 & 39.90 & 86.45 & 79.36 & 53.02 & 87.88 & 78.22 & 58.34 & 80.42 & 56.25 \\
Qwen-3-0.6B          & 65.02 & 77.59 & 93.19 & 71.59 & 86.24 & 91.05 & 79.91 & 85.96 & 91.85 & 59.38 & 91.67 & 91.67 \\
\bottomrule
\end{tabular}}
\caption{
Full breakdown of \textbf{Keypoint Coverage (\%)} across four difficulty levels under \textbf{Base}, \textbf{Mix}, and \textbf{Golden} retrieval settings for all 20 models. 
Higher coverage indicates better factual completeness.}
\label{tab:difficulty_full}
\end{table*}

\section{General-Domain Comparison (MMLU-Pro Subset Evaluation)}
\label{appendix:mmlu_results}

To contextualize the domain specificity of DisastQA, 
we further evaluated the same set of 20 models on an \textbf{8-subject subset} of \textbf{MMLU-Pro}~\cite{wang2024mmlu-pro}.  
This subset spans the subjects \textit{Biology, Psychology, Economics, Business, Engineering, Chemistry, Law}, and \textit{Health}, 
covering a total of 2,000 multiple-choice questions (each with ten options) under a \textit{0-shot} evaluation setting.

Table~\ref{tab:mmlu_subset} summarizes the average accuracies and ranks across all evaluated models.
These results serve as a \textbf{general-domain baseline} for comparison with 
\textbf{DisastQA-MCQ (Golden)}, as visualized in Figure~\ref{fig:mcq_performance}.  
While absolute accuracies are not directly comparable (MMLU-Pro is a closed-book benchmark, 
whereas DisastQA includes gold passages), 
the table provides a clear reference for understanding 
cross-domain ranking divergences discussed in Section~\ref{sec:results}.

\section{Representative Error Case Analysis}
\label{appendix:error_cases}

To better understand the limitations of LLMs on DisastQA, 
we provide representative examples of common failure patterns observed across both MCQ and OE settings.  
This qualitative inspection complements the quantitative results in Section~\ref{sec:results}.  
Errors typically fall into two categories: (1) \textit{distractor confusion}, where models are misled by numerically or semantically similar distractors, 
and (2) \textit{incomplete factual coverage}, where generated answers fail to capture all keypoints from the reference.  
Representative examples of these two failure types are shown in Table~\ref{tab:error_cases}.

\begin{table*}[t]
\centering
\small
\setlength{\tabcolsep}{4pt}
\renewcommand{\arraystretch}{1.1}
\begin{tabular}{p{0.95\textwidth}}
\toprule
\textbf{(a) MCQ and OE Failure Cases — Example Set 1} \\
\midrule
\textbf{MCQ:} What was the reported magnitude of the earthquake in City~X? \\
\textbf{Passage:} The earthquake struck City~X on March 12, 2020, with a magnitude of 6.7, causing severe building damage. \\
\textbf{Options:} (A) 5.7 \quad (B) 6.3 \quad (C) \textbf{6.7} \quad (D) 7.2 \\
\textbf{Prediction:} (D) 7.2 \\
\textbf{Error Type:} Distractor-based confusion (numerical alteration). \\
\midrule
\textbf{OE:} What were the main impacts of Hurricane~Y? \\
\textbf{Passage:} Hurricane~Y caused widespread flooding across coastal towns, disrupted power supply to over 200{,}000 households, and led to multiple road closures. \\
\textbf{Reference Answer:} Flooding in coastal towns; power outage affecting 200{,}000 households; road closures. \\
\textbf{Model Answer:} Hurricane~Y caused flooding and severe damage in coastal towns. \\
\textbf{Coverage:} 1/3 keypoints (only flooding captured; missed power outage and road closures). \\
\textbf{Error Type:} Incomplete factual coverage. \\
\midrule
\textbf{(b) MCQ and OE Failure Cases — Example Set 2} \\
\midrule
\textbf{MCQ:} What was the recorded peak flood depth in River~Z’s low-lying district? \\
\textbf{Passage:} Field reports indicate a peak flood depth of 1.5\,m near the River~Z levee breach, forcing evacuations in adjacent neighborhoods. \\
\textbf{Options:} (A) 0.5\,m \quad (B) 1.0\,m \quad (C) \textbf{1.5\,m} \quad (D) 2.0\,m \\
\textbf{Prediction:} (B) 1.0\,m \\
\textbf{Error Type:} Underestimation of numeric fact (ignored explicit value in passage). \\
\midrule
\textbf{OE:} Summarize the key impacts of the Wildfire~K incident. \\
\textbf{Passage:} Wildfire~K burned 48{,}000 acres, prompted evacuations of 15{,}000 residents, and pushed local AQI to ``hazardous'' for three consecutive days. \\
\textbf{Reference Answer:} 48k acres burned; 15k residents evacuated; AQI ``hazardous'' for 3 days. \\
\textbf{Model Answer:} Wildfire~K forced large-scale evacuations in nearby communities. \\
\textbf{Coverage:} 1/3 keypoints (mentions evacuations; misses acreage and AQI duration). \\
\textbf{Error Type:} Missing quantitative details and duration qualifiers. \\
\bottomrule
\end{tabular}
\caption{
Representative MCQ and OE failure cases from \textbf{DisastQA}.  
Example~(a) shows numerical distractor confusion and incomplete factual coverage, 
while Example~(b) highlights quantitative underestimation and omission of temporal keypoints—illustrating challenges in precise factual grounding despite fluent generation.
}
\label{tab:error_cases}
\end{table*}

\subsection{Comparison of LLM Drafts and Human-Refined Versions}
\label{app:error_refine_compare}

To further demonstrate how human verification enhances factual grounding and clarity, 
we provide representative cases contrasting raw LLM drafts with their human-refined counterparts.
These examples illustrate the most frequent error patterns corrected during refinement— 
including vague question framing, factual omission, and implausible distractors.  
Table~\ref{tab:error_refinement} presents four paired examples (two MCQ and two OE) showing how human experts ensured precision and passage alignment.

\begin{table*}[t]
\centering
\small
\setlength{\tabcolsep}{5pt}
\renewcommand{\arraystretch}{1.1}
\resizebox{\textwidth}{!}{
\begin{tabular}{p{0.47\textwidth} | p{0.47\textwidth}}
\toprule
\textbf{LLM Draft} & \textbf{Human-Refined Version} \\
\midrule
\multicolumn{2}{l}{\textbf{MCQ Example 1 (Medical Device Safety)}} \\
\midrule
\textbf{Question:} How should you handle a medical test device? \newline
\textbf{Passage:} The document is a Safety Data Sheet (SDS) for the Proflow™ Hepatitis A Test Device and Sample Preparation Device. Industries must actively engage in public campaigns to inform communities about POPs (Persistent Organic Pollutants) and their associated risks. \newline
\textbf{Options:} (A) Keep it in sunlight \quad (B) Use protection gear \quad (C) Throw in regular trash \quad (D) Only use outside \newline
\textbf{Issue:} Question too vague; options unclear about specific safety requirements. &
\textbf{Question:} What is one important precaution to follow when handling the Proflow™ Hepatitis A Test Device? \newline
\textbf{Passage:} Same as left. \newline
\textbf{Options:} (A) Store the device in direct sunlight to maintain its efficacy \quad (B) \textbf{Ensure proper ventilation and wear personal protective equipment during use} \quad (C) Dispose of the device in regular trash without special handling precautions \quad (D) Only use the device outdoors to avoid contamination \newline
\textbf{Fix:} Clarifies specific device and precise safety requirements; options clearly distinguish correct safety practices. \\
\midrule
\multicolumn{2}{l}{\textbf{MCQ Example 2 (Environmental Impact)}} \\
\midrule
\textbf{Question:} What causes acid rain problems? \newline
\textbf{Passage:} The burning of fossil fuels in power plants and vehicles contributes significantly to acid rain. This process releases sulfur dioxide and nitrogen oxides into the atmosphere, which react with water vapor to form acidic compounds. \newline
\textbf{Options:} (A) Natural processes \quad (B) Burning coal and oil \quad (C) Water pollution \quad (D) Industrial waste \newline
\textbf{Issue:} Question too broad; options mix different pollution types; correct answer not specific enough. &
\textbf{Question:} What are the primary human activities that lead to the formation of acid rain? \newline
\textbf{Passage:} Same as left. \newline
\textbf{Options:} (A) Natural volcanic emissions \quad (B) \textbf{Burning fossil fuels in power plants and vehicles} \quad (C) Agricultural fertilizer use \quad (D) Municipal waste disposal \newline
\textbf{Fix:} Specifies “human activities” and “formation”; options clearly distinguish between natural and anthropogenic sources. \\
\midrule
\multicolumn{2}{l}{\textbf{OE Example 1 (Electronic Waste)}} \\
\midrule
\textbf{Question:} What causes electronic waste in Indian cities? \newline
\textbf{Passage:} The rapid expansion of the IT industry in Bangalore is emblematic of a larger trend in many of India's urban centers. The predominant contributors to the electronic waste problem are classified into three main categories: large household appliances, which account for 42\%, followed by information and communication technology (ICT) equipment at 34\%, and consumer electronics making up the remaining 14\%. \newline
\textbf{Model Draft Answer:} IT growth causes more electronic waste and environmental problems. \newline
\textbf{Issue:} Over-general; misses specific categories and percentages; lacks quantitative details. &
\textbf{Question:} What are the main categories contributing to the electronic waste problem in Indian cities experiencing rapid IT industry growth, and what percentage does each category represent? \newline
\textbf{Passage:} Same as left. \newline
\textbf{Human Reference Answer:} The main categories are: large household appliances (42\%), information and communication technology equipment (34\%), and consumer electronics (14\%). \newline
\textbf{Fix:} Specifies exact categories with precise percentages; addresses both parts of the question comprehensively. \\
\midrule
\multicolumn{2}{l}{\textbf{OE Example 2 (Energy Technology)}} \\
\midrule
\textbf{Question:} How can we reduce energy environmental impacts? \newline
\textbf{Passage:} In light of the significant environmental impacts posed by non-combusted substances and the limitations of current combustion efficiencies, adopting cleaner and more efficient combustion technologies, alongside a transition to renewable energy sources, is imperative for lasting environmental protection. \newline
\textbf{Model Draft Answer:} Use better energy methods and cleaner options to help the environment. \newline
\textbf{Issue:} Over-general; misses “cleaner combustion” and “renewables transition” as explicit measures; vague phrasing. &
\textbf{Question:} What concrete measures does the passage recommend to mitigate energy-related environmental impacts? \newline
\textbf{Passage:} Same as left. \newline
\textbf{Human Reference Answer:} The passage suggests adopting cleaner and more efficient combustion technologies and transitioning to renewable energy sources for reducing the environmental impact. \newline
\textbf{Fix:} Specifies both required measures explicitly; removes vague phrasing; aligns precisely with passage content. \\
\bottomrule
\end{tabular}}
\caption{
Representative MCQ and OE refinement examples from \textbf{DisastQA}.  
Human refinement clarifies conceptual scope, corrects factual omissions, and ensures full alignment with passage-grounded evidence.
}
\label{tab:error_refinement}
\end{table*}

\subsection{Failure Mode Distribution}
\label{app:failure_distribution}

Table~\ref{tab:failure_type_dist} summarizes the approximate distribution of common LLM draft failure types observed during the refinement phase, 
complementing the qualitative examples above (Table~\ref{tab:error_cases} and Table~\ref{tab:error_refinement}).  

\begin{table}[h]
\centering
\small
\setlength{\tabcolsep}{6pt}
\renewcommand{\arraystretch}{1.1}
\begin{tabular}{lrr}
\toprule
\textbf{Failure Type} & \textbf{MCQ (\%)} & \textbf{OE (\%)} \\
\midrule
Implausible / trivial distractors & 35\% & -- \\
Factual inaccuracies in stems & 20\% & -- \\
Ambiguous wording & 15\% & 20\% \\
Incomplete coverage & -- & 25\% \\
Overly generic answers & -- & 15\% \\
Other minor issues & 5\% & 15\% \\
\bottomrule
\end{tabular}
\caption{
Distribution of common LLM draft issues observed during human refinement. 
Percentages are approximate, based on annotation review logs. 
While most drafts were structurally valid, human intervention remained essential for improving clarity, factual grounding, and coverage.
}
\label{tab:failure_type_dist}
\end{table}

Finally, Table~\ref{tab:filter_stats} details the filtering process from the DisastIR corpus to the finalized benchmark tasks, 
and Table~\ref{tab:oe_keypoints_stats} reports token-level statistics of annotated keypoints, 
which quantify factual density within the OE evaluation set.

\begin{table}[t]
\centering
\small
\setlength{\tabcolsep}{3.5pt}
\renewcommand{\arraystretch}{1.05}
\resizebox{\linewidth}{!}{%
\begin{tabular}{lcc}
\toprule
\textbf{Stage} & \textbf{\#Items} & \textbf{Notes} \\
\midrule
Candidates (score = 3) & 5,740 & Retrieved from DisastIR \\
After stratified sampling & 3,000 & Balanced over intent × event \\
MCQ verified & 2,000 & Human-refined, solve-checked \\
OE verified & 1,000 & Human-refined, gold answers authored \\
OE keypoint subset & 200 & Double-pass consensus verified \\
\bottomrule
\end{tabular}}
\caption{
Filtering process from the DisastIR corpus to the final DisastQA tasks.
}
\label{tab:filter_stats}
\end{table}

\paragraph{Keypoint Statistics.}
\begin{table}[t]
\centering
\small
\begin{tabular}{ccccc}
\toprule
\textbf{\#Keypoints} & \textbf{Count} & \textbf{Mean Tokens} & \textbf{Min} & \textbf{Max} \\
\midrule
1  & 5   & 15.4  & 11  & 24  \\
2  & 104 & 38.1  & 12  & 95  \\
3  & 181 & 65.4  & 16  & 232 \\
4  & 244 & 97.5  & 22  & 248 \\
5  & 259 & 123.6 & 34  & 298 \\
6  & 129 & 144.2 & 42  & 291 \\
7  & 52  & 156.0 & 86  & 289 \\
8  & 15  & 182.3 & 136 & 264 \\
9  & 6   & 215.5 & 67  & 350 \\
10 & 3   & 188.7 & 129 & 250 \\
11 & 0   & --    & --  & --  \\
12 & 1   & 209.0 & 209 & 209 \\
13 & 0   & --    & --  & --  \\
14 & 1   & 240.0 & 240 & 240 \\
\bottomrule
\end{tabular}
\caption{Answer length statistics by the number of annotated keypoints in the OE portion of DisastQA. 
Mean, Min, and Max are measured in tokens. Keypoint counts 11 and 13 do not occur in the dataset.}
\label{tab:oe_keypoints_stats}
\end{table}

These examples and statistics collectively confirm that while LLMs accelerate item drafting, 
human refinement remains indispensable for factual grounding, plausibility, and completeness in safety-critical QA benchmarks like \textbf{DisastQA}.

\section{Construction Algorithm}
\label{appendix:algo}

\begin{algorithm}[h]
\caption{DisastQA Construction Pipeline (Human--LLM Collaboration)}
\label{alg:qa_construction}
\begin{algorithmic}[1]
\Require Source corpus $\mathcal{D}$ (queries $q$, passages $p_g$)
\Ensure Verified MCQ set $\mathcal{Q}_{MCQ}$, Verified OE set $\mathcal{Q}_{OE}$

\State Sample $(q, p_g)$ pairs from $\mathcal{D}$ based on event type
\State Assign each pair to MCQ or OE task

\For{each $(q, p_g)$ with assigned task}
    \State \textbf{Rewrite:} LLM transforms query $q$ into question $Q$ using $p_g$
    
    \If{task = MCQ}
        \State LLM generates gold answer and 3 distractors
        \State Human verifies correctness and performs solve-check
        \State Add to $\mathcal{Q}_{MCQ}$
    \Else \Comment{OE task}
        \State LLM generates reference answer $A$ grounded in $p_g$
        \State Human refines $A$ for factual completeness
        \If{in keypoint annotation subset}
            \State Human annotates atomic keypoints
        \EndIf
        \State Add to $\mathcal{Q}_{OE}$
    \EndIf
\EndFor
\end{algorithmic}
\end{algorithm}

\subsection{Filtering Process Statistics}
\label{appendix:filter_stats}

Table~\ref{tab:filter_stats} summarizes the stepwise filtering pipeline 
from the DisastIR candidate pool to the finalized DisastQA tasks.
The process ensures balanced sampling across intent–event combinations, 
followed by human verification for both MCQ and OE items.

\subsection{Annotation Details}
The following sections expand upon the annotation protocol introduced in Section~\ref{sec:mcq_construction} and ~\ref{sec:oe_construction}, 
providing detailed task-specific guidelines for MCQ and OE verification.

\subsubsection{General Annotation Protocol}
\label{appendix:protocol}

All annotation followed a consensus-based protocol to ensure factual accuracy and consistency across the dataset.
\begin{itemize}[noitemsep,leftmargin=*]
    \item \textbf{Annotators:} one senior Ph.D. researcher and three Ph.D. students specializing in disaster information analysis.  
    \item \textbf{Process:} independent annotation $\rightarrow$ second-pass review $\rightarrow$ group discussion $\rightarrow$ consensus verification.

    \item \textbf{Division of labor:} MCQ items were verified and refined primarily by the Ph.D. students, while OE questions and reference answers were reviewed and finalized by the senior researcher.  
\end{itemize}

\subsubsection{Task-specific Annotation Guidelines for MCQ and OE}
\label{appendix:guidelines}

\paragraph{MCQ Annotation.}
\begin{itemize}[noitemsep,leftmargin=*]
    \item Rewrite drafts that are ambiguous, ungrounded, or under-specified.
    \item Ensure that the correct answer is fully supported by the reference passage $p_g$.
    \item Design distractors to be plausible yet unsupported by $p_g$; avoid trivial negations or irrelevant options.

See Table~\ref{tab:mmlu_subset} for model performance on the MMLU-Pro baseline used for cross-domain comparison.
This table complements Figure~\ref{fig:mcq_performance} in the main text.

\end{itemize}
\begin{table}[h]
\centering
\small
\setlength{\tabcolsep}{5pt}
\renewcommand{\arraystretch}{1.05}
\begin{tabular}{lcc}
\toprule
\textbf{Model} & \textbf{Accuracy (\%)} & \textbf{Rank} \\
\midrule
GPT-5.2 & 65.2 & 1 \\
Gemini-3-Pro & 50.0 & 2 \\
Qwen-3-8B & 47.3 & 3 \\
Qwen-3-4B & 41.6 & 4 \\
GPT-4o & 41.5 & 5 \\
Gemini-1.5-Pro & 40.5 & 6 \\
Llama-3-8B & 39.9 & 7 \\
Yi-6B-Chat & 33.6 & 8 \\
Llama-3.2-3B-Instruct & 31.9 & 9 \\
Mistral-7B-Instruct-v0.2 & 31.3 & 10 \\
Hunyuan-4B-Instruct & 28.6 & 11 \\
Hunyuan-7B-Instruct & 28.3 & 12 \\
Qwen-2.5-3B-Instruct & 27.1 & 13 \\
DeepSeek-v3-7B & 25.9 & 14 \\
Gemma-7B & 25.7 & 15 \\
Phi-2 & 23.9 & 16 \\
AceMath-1.5B-Instruct & 22.5 & 17 \\
Llama-3.2-1B-Instruct & 21.0 & 18 \\
Qwen-3-0.6B & 17.3 & 19 \\
Falcon-3-1B-Instruct & 10.6 & 20 \\
\bottomrule
\end{tabular}
\caption{
Performance of 20 models on the \textbf{8-subject MMLU-Pro subset} (2,000 questions, 0-shot). 
This subset is used as a general-domain baseline for cross-domain ranking comparison in Figure~\ref{fig:mcq_performance}.
}
\label{tab:mmlu_subset}
\end{table}
\paragraph{OE Annotation.}
\begin{itemize}[noitemsep,leftmargin=*]
    \item Rewrite drafts that are unclear or not strictly answerable from $p_g$.
    \item Author all reference answers manually to ensure conciseness and factual grounding.
    \item Decompose each reference answer into keypoints, where each keypoint represents one minimal factual unit. 
          The number of keypoints varies with the complexity of the answer.
\end{itemize}

These keypoints form the basis for the human-verified Keypoint Coverage
metric defined in Section~\ref{sec:oe_eval}.

\paragraph{Example.}
An example of human refinement illustrating conceptual improvement and distractor quality enhancement is shown in Table~\ref{tab:mcq_refinement_example}.

\begin{table*}[t]
\centering
\small
\setlength{\tabcolsep}{4pt}
\renewcommand{\arraystretch}{1.1}
\begin{tabular}{p{0.3\textwidth} p{0.3\textwidth} p{0.3\textwidth}}
\toprule
\textbf{LLM Draft} & \textbf{Human Refined} & \textbf{Revision Rationale} \\
\midrule
Q: Which policy supports floodplain management? &
Q: Which policy best embodies the ``No Adverse Impact'' principle in floodplain management? &
Draft vague; refined to target the named principle and ensure grounding in the passage’s definition of NAI (no increased risk, cumulative-impact review). \\
\midrule
Distractor: ``Approve variances when projects boost tax revenue'' &
Distractor: ``Allow fill in floodways if compensatory storage is added off-site'' &
Original distractor is obviously off-policy; refined to a realistic but unsupported alternative often confused with NAI, improving plausibility without contradicting the passage. \\
\bottomrule
\end{tabular}
\caption{Illustrative MCQ refinement demonstrating improvements in conceptual grounding and distractor plausibility 
for floodplain management under the ``No Adverse Impact'' (NAI) principle.
}
\label{tab:mcq_refinement_example}
\end{table*}

\subsection{Keypoint Annotation Examples}
\label{appendix:keypoint_example}

As introduced in Section~\ref{sec:oe_eval}, each human-authored reference answer is decomposed into minimal, independently verifiable factual units called \textit{keypoints}. 
During evaluation, model-generated answers are scored by the proportion of gold keypoints they correctly cover, as defined by Equation~\ref{eq:coverage} in the main text. 
Representative decompositions across multiple difficulty levels are summarized in Table~\ref{tab:keypoint_examples}, 
showing how reference answers are broken down into fine-grained factual components that capture factual completeness beyond surface overlap.

\begin{table*}[t]
\centering
\small
\setlength{\tabcolsep}{4pt}
\renewcommand{\arraystretch}{1.05}
\resizebox{\textwidth}{!}{%
\begin{tabular}{p{0.09\textwidth} p{0.45\textwidth} p{0.42\textwidth}}
\toprule
\textbf{Difficulty / KP} & \textbf{Question \& Golden Answer (excerpt)} & \textbf{Human-Annotated Keypoints} \\
\midrule
\textbf{Easy (2)} &
\textit{Q:} What measures does the passage suggest for reducing the environmental impact of energy-related activities? \newline
\textit{A:} The passage suggests adopting cleaner and more efficient combustion technologies and transitioning to renewable energy sources. &
\textbf{(1)} Adopting cleaner and more efficient combustion technologies. \newline
\textbf{(2)} Transitioning to renewable energy sources. \\

\midrule
\textbf{Medium (3)} &
\textit{Q:} What are the key responsibilities and actions that building owners must undertake after ACM cladding is identified? \newline
\textit{A:} Building owners must implement interim safety advice to ensure resident safety; measures reduce risk temporarily until remediation; and local authorities coordinate with fire services to enforce safety. &
\textbf{(1)} Implement interim safety advice for resident safety. \newline
\textbf{(2)} Interim measures reduce risk until remediation. \newline
\textbf{(3)} Local authorities and fire services coordinate with owners. \\

\midrule
\textbf{Medium (5)} &
\textit{Q:} What recent advancements have emerged in studies on suspension bridges during fire incidents? \newline
\textit{A:} Developed a numerical model using genetic algorithms to simulate cable behavior under fire; analyzed cooling with finite element methods; emphasized risks and maintenance strategies. &
\textbf{(1)} Developed a numerical model for high-temperature cable behavior. \newline
\textbf{(2)} Used Genetic Algorithms to optimize stiffness parameters. \newline
\textbf{(3)} Highlighted hydrocarbon fire risks to bridge safety. \newline
\textbf{(4)} Conducted finite element simulations for cooling. \newline
\textbf{(5)} Provided recommendations for engineering practice. \\

\midrule
\textbf{Hard (6)} &
\textit{Q:} How is effective decontamination achieved for all effluent within containment perimeters? \newline
\textit{A:} Achieved through validated inactivation procedures; includes pass-through autoclave, HEPA filtration, cycle alarms, gaseous decontamination, and dunk tank with active compound. &
\textbf{(1)} Validated inactivation procedure. \newline
\textbf{(2)} Pass-through autoclave with bioseal/interlocks. \newline
\textbf{(3)} HEPA filtration of air discharge. \newline
\textbf{(4)} Cycle recording/alarm systems. \newline
\textbf{(5)} Gaseous decontamination chamber. \newline
\textbf{(6)} Dunk tank with active compound. \\

\midrule
\textbf{Extremely Complex (8)} &
\textit{Q:} How can machine learning enhance thunderstorm hazard predictions through integration of radar, satellite, and lightning data? \newline
\textit{A:} Radar data most predictive; satellite aids where radar is sparse; lightning data valuable for specific hazards; source selection affects model accuracy and overfitting. &
\textbf{(1)} Integrate radar, satellite, lightning, and NWP data. \newline
\textbf{(2)} Radar is the most predictive for storm indicators. \newline
\textbf{(3)} Satellite imagery improves prediction where radar is sparse. \newline
\textbf{(4)} Lightning data helps forecast lightning-specific hazards. \newline
\textbf{(5)} Source selection impacts accuracy. \newline
\textbf{(6)} Avoid overfitting via careful data selection. \newline
\textbf{(7)} Understand contributions of individual data sources. \newline
\textbf{(8)} Explore additional ML methods for improvement. \\
\bottomrule
\end{tabular}}%
\caption{
Representative keypoint annotation examples across difficulty levels. 
Each example shows the human-authored gold answer and its decomposition into atomic factual units (keypoints), 
which form the basis for factual completeness evaluation in DisastQA-OE.
}
\label{tab:keypoint_examples}
\end{table*}

\vspace{1ex}
\noindent
These structured examples complement the quantitative statistics in Appendix~\ref{tab:oe_keypoints_stats} and demonstrate how the keypoint-based framework operationalizes factual completeness across question complexities.

\subsection{Prompt Templates}
\label{appendix:prompt}

Prompt templates follow the general generation procedure outlined in Section~\ref{sec:evaluation_methodology}, 
where both Multiple-Choice (MCQ) and Open-Ended (OE) items were constructed through structured prompts grounded in retrieved passages. 
These templates operationalize the Base, Mix, and Golden evaluation contexts described in the main text. Prompts are provided for transparency and reproducibility, and were not tuned
for individual models.

\paragraph{MCQ Generation.}
The following real instance demonstrates how an MCQ was generated 
from an actual reference passage in the \textbf{DisastQA} dataset 
(see Appendix~\ref{app:prompts}).

\begin{table*}[t]
\centering
\small
\begin{tabular}{p{0.9\textwidth}}
\toprule
\textbf{User Query:}\par
Enhance floodplain management under the National Flood Insurance Program (NFIP).

\medskip
\textbf{Reference Passage (excerpt):}\par
Incorporating the ASFPM's ``No Adverse Impact'' policy into local floodplain management programs.
Revising the floodplain ordinance to incorporate cumulative substantial damage requirements.
Adopting a ``no--rise'' in base flood elevation clause for the flood damage prevention ordinance.
Extending the freeboard requirement beyond the mapped floodplain to include equivalent land elevation.
Including requirements in the local floodplain ordinance for homeowners to sign non-conversion agreements for areas below base flood elevation.
Developing an educational flyer for NFIP policyholders on increased cost of compliance during post-flood damage assessments.
Annually notifying owners of repetitive loss properties of Flood Mitigation Assistance funding.
Offering incentives for building above the required freeboard minimum (``code plus'').

\medskip
\textbf{Generated MCQ:}\par
{\ttfamily
Question: Which of the following practices should local governments implement to enhance their floodplain management under the National Flood Insurance Program?\par
Options:\par
A. Establish a program to subsidize insurance for all property owners regardless of flood risk\par
B. Implement a mandatory evacuation plan for all residents during flood warnings\par
C. Revise the floodplain ordinance to include cumulative substantial damage requirements\par
D. Provide tax breaks for property development in flood-prone areas\par
Correct Option: C\par
Reason: The passage explicitly lists ``revising the floodplain ordinance to include cumulative substantial damage requirements'' as a recommended practice.
}
\\
\bottomrule
\end{tabular}
\caption{
Real MCQ generation example from \textbf{DisastQA}.
This instance shows how the generation prompt creates a question grounded in the domain passage, with realistic distractors and clear factual justification for the correct answer.
}
\label{tab:mcq_prompt}
\end{table*}

\paragraph{OE Generation.}
The OE prompt follows the same structure as the MCQ template but omits options and the correct answer, producing only the question text grounded in the passage.  
These templates correspond to the data generation settings used in Section~\ref{sec:evaluation_methodology} (\textit{Base}, \textit{Mix}, and \textit{Golden} contexts).

\paragraph{Mix Context Construction.}
To simulate realistic noisy retrieval, each question in the Mix setting is paired with one gold passage ($p_g$) and four distractor passages ($p_d$) sampled from the DisastIR corpus across different relevance levels (scores 0–3). 
Specifically, we include one high-relevance distractor (score = 3 but non-gold), one medium (score = 2), one weak (score = 1), and one irrelevant (score = 0) passage whenever available. 
This design preserves the gold evidence while introducing both semantically similar and noisy contexts, approximating imperfect retrieval conditions where correct and misleading information coexist. 
These Mix contexts follow the tri-level evaluation setup described in Section~\ref{sec:evaluation_methodology}.

\section{Model Implementation Details}
\label{appendix:model_details}

Table~\ref{tab:models} summarizes all evaluated models, including model sizes, 
Hugging Face identifiers or API endpoints, licenses, and public access links. 
All open-source models were loaded via \texttt{transformers} in half-precision (fp16/bf16) mode. 
Closed-source models (\textit{GPT}, \textit{Gemini}) were accessed through their official APIs.

\begin{table*}[t]
\centering
\small
\setlength{\tabcolsep}{4pt}
\renewcommand{\arraystretch}{1.05}
\begin{tabular}{p{0.16\linewidth} p{0.08\linewidth} p{0.40\linewidth} p{0.14\linewidth} p{0.10\linewidth}}
\toprule
\textbf{Model} & \textbf{Params} & \textbf{Identifier / Endpoint} & \textbf{License} & \textbf{Link} \\
\midrule
Qwen3-0.6B & 0.75B & Qwen/Qwen3-0.6B & Apache 2.0 & \href{https://huggingface.co/Qwen/Qwen3-0.6B}{HF} \\
Qwen3-4B & 4.02B & Qwen/Qwen3-4B & Apache 2.0 & \href{https://huggingface.co/Qwen/Qwen3-4B}{HF} \\
Qwen3-8B & 8.19B & Qwen/Qwen3-8B & Apache 2.0 & \href{https://huggingface.co/Qwen/Qwen3-8B}{HF} \\
Llama-3.2-1B & 1.24B & meta-llama/Llama-3.2-1B-Instruct & Meta Llama 3.2 & \href{https://huggingface.co/meta-llama/Llama-3.2-1B-Instruct}{HF} \\
Llama-3.2-3B & 3.21B & meta-llama/Llama-3.2-3B-Instruct & Meta Llama 3.2 & \href{https://huggingface.co/meta-llama/Llama-3.2-3B-Instruct}{HF} \\
Llama-3-8B & 8.03B & meta-llama/Meta-Llama-3-8B-Instruct & Meta Llama 3 & \href{https://huggingface.co/meta-llama/Meta-Llama-3-8B-Instruct}{HF} \\
Gemma-7B & 8.54B & google/gemma-7b-it & Gemma License & \href{https://huggingface.co/google/gemma-7b-it}{HF} \\
Mistral-7B & 7.24B & mistralai/Mistral-7B-Instruct-v0.2 & Apache 2.0 & \href{https://huggingface.co/mistralai/Mistral-7B-Instruct-v0.2}{HF} \\
Phi-2 & 2.78B & microsoft/phi-2 & MIT & \href{https://huggingface.co/microsoft/phi-2}{HF} \\
Falcon-3-1B & 1.67B & tiiuae/Falcon3-1B-Instruct & Apache 2.0 & \href{https://huggingface.co/tiiuae/Falcon3-1B-Instruct}{HF} \\
Yi-6B & 6.06B & 01-ai/Yi-6B-Chat & Apache 2.0 & \href{https://huggingface.co/01-ai/Yi-6B-Chat}{HF} \\
DeepSeek-7B & $\sim$7B & deepseek-ai/deepseek-llm-7b-chat & Apache 2.0 & \href{https://huggingface.co/deepseek-ai/deepseek-llm-7b-chat}{HF} \\
Hunyuan-4B & 4.22B & tencent/Hunyuan-4B-Instruct & Apache 2.0 & \href{https://huggingface.co/tencent/Hunyuan-4B-Instruct}{HF} \\
Hunyuan-7B & 7.50B & tencent/Hunyuan-7B-Instruct & Apache 2.0 & \href{https://huggingface.co/tencent/Hunyuan-7B-Instruct}{HF} \\
AceMath-1.5B & 1.78B & nvidia/AceMath-1.5B-Instruct & Apache 2.0 & \href{https://huggingface.co/nvidia/AceMath-1.5B-Instruct}{HF} \\
\midrule
GPT-5.2 & N/A & OpenAI API (gpt-5.2) & Proprietary & \href{https://openai.com/product}{API} \\
Gemini-3 Pro & N/A & Google API (gemini-3-pro) & Proprietary & \href{https://ai.google.dev/models/gemini}{API} \\
GPT-4o & N/A & OpenAI API (gpt-4o) & Proprietary & \href{https://openai.com/product/gpt-4o}{API} \\
Gemini-1.5 Pro & N/A & Google API (gemini-1.5-pro) & Proprietary & \href{https://ai.google.dev/models/gemini}{API} \\
\bottomrule
\end{tabular}
\caption{
Evaluated models with parameter counts, identifiers, and license terms.  
"HF" denotes public Hugging Face repositories; full URLs are provided in the project's supplementary README.  
Note: Qwen3-0.6B corresponds to a 752M-parameter model.
}
\label{tab:models}
\end{table*}

\section{Per-model Difficulty Analysis}
\label{appendix:difficulty_results}

For completeness, we report detailed \textbf{per-model Keypoint Coverage results}
across four difficulty levels (\textbf{Easy}, \textbf{Medium}, \textbf{Hard}, and \textbf{Extremely Complex})
under three retrieval settings (\textbf{Base}, \textbf{Mix}, and \textbf{Golden}).  
These results complement the overall difficulty trends discussed in Section~\ref{sec:results}:
retrieval consistently improves factual coverage across all difficulty levels, 
yet performance still declines as the number of required keypoints increases.

Table~\ref{tab:coverage_difficulty_models_all} presents the full per-model results 
for all 20 evaluated models. The table is organized to facilitate side-by-side comparisons 
across model families, highlighting the consistent gains brought by Mix and Golden retrievals 
as well as the performance degradation for more complex multi-fact questions.

\section{Prompt Templates}
\label{app:prompts}

We provide the standardized prompts used for both 
\textbf{Multiple-Choice Questions (MCQ)} and 
\textbf{Open-Ended Questions (OE)} across the three evaluation settings:
\textbf{Base}, \textbf{Mix}, and \textbf{Golden}.

\subsection{Multiple-Choice Questions (MCQ)}

\textbf{Mix Setting (1 Golden + 4 Distractors).}  
This is the most challenging configuration, requiring the model to identify correct evidence among distractors.

\begin{tcolorbox}[colback=gray!3,colframe=gray!60,breakable]
\textbf{MCQ — Mix Prompt}

You are given 5 passages retrieved from a disaster information corpus.
Some passages may be irrelevant or only partially related.
Read all passages carefully and answer the question based on the most relevant and factually correct passage(s).

Write the passage used: \texttt{Passage: <1–5>}  
Then write the answer: \texttt{Answer: <A–D>}

\textbf{Passage 1:} \{p1\} \\
\textbf{Passage 2:} \{p2\} \\
\textbf{Passage 3:} \{p3\} \\
\textbf{Passage 4:} \{p4\} \\
\textbf{Passage 5:} \{p5\}

\textbf{Question:} \{q\}  
\textbf{Options:} \{opts\}

Answer:
\end{tcolorbox}

\textbf{Golden Setting (1 Golden Passage).}

\begin{tcolorbox}[colback=gray!3,colframe=gray!60,breakable]
\textbf{MCQ — Golden Prompt}

\textbf{Passage:} \{gold\_passage\}  
\textbf{Question:} \{q\}  
\textbf{Options:} \{opts\}

Answer only the letter (A--D).  
Answer:
\end{tcolorbox}

\textbf{Base Setting (No Passage).}

\begin{tcolorbox}[colback=gray!3,colframe=gray!60,breakable]
\textbf{MCQ — Base Prompt}

\textbf{Question:} \{q\}  
\textbf{Options:} \{opts\}

Answer only the letter (A--D).  
Answer:
\end{tcolorbox}

\subsection{Open-Ended Questions (OE)}
\textbf{Mix Setting (1 Golden + 4 Distractors).}
This setting evaluates the model's ability to discriminate relevant evidence from noise and generate accurate responses. 
Crucially, the model is explicitly instructed to identify the source passage before generating the answer, enhancing interpretability.

\begin{tcolorbox}[colback=gray!3,colframe=gray!60,breakable]
\textbf{OE — Mix Prompt}

You are given 5 passages (some may be irrelevant). You must select ONLY ONE passage that is most relevant to answer the question. First output the passage number you selected in the format \texttt{Passage: <single number between 1 and 5>} (only one number, no commas or multiple numbers), then provide a comprehensive answer based on that passage.

\textbf{Passage 1:} \{p1\} \\
\textbf{Passage 2:} \{p2\} \\
\textbf{Passage 3:} \{p3\} \\
\textbf{Passage 4:} \{p4\} \\
\textbf{Passage 5:} \{p5\}

\textbf{Question:} \{query\} \\
\textbf{Difficulty Level:} \{difficulty\}

\textbf{Instructions:} Provide a comprehensive answer within \{word\_limit\} words. Ensure you cover all important aspects and key points related to the question. Be thorough but concise.

\textbf{Answer:}

\tcblower
\footnotesize \textit{\textbf{Implementation Note:} The \texttt{\{word\_limit\}} variable is dynamically set (e.g., $\approx$50 words for Easy, $\approx$150 for Hard) to enforce information density and prevent length-based gaming of coverage metrics.}
\end{tcolorbox}

\textbf{Golden Setting (1 Golden Passage).}

\begin{tcolorbox}[colback=gray!3,colframe=gray!60,breakable]
\textbf{OE — Golden Prompt}

\textbf{Passage:} \{gold\_passage\} \\
\textbf{Question:} \{query\} \\
\textbf{Difficulty Level:} \{difficulty\}

\textbf{Instructions:} Provide a comprehensive answer within \{word\_limit\} words. Ensure you cover all important aspects and key points related to the question. Be thorough but concise.

\textbf{Answer:}

\tcblower
\footnotesize \textit{\textbf{Note:} The same adaptive length constraint applies here to ensure fair comparison.}
\end{tcolorbox}

\textbf{Base Setting (No Passage).}

\begin{tcolorbox}[colback=gray!3,colframe=gray!60,breakable]
\textbf{OE — Base Prompt}

\textbf{Question:} \{query\} \\
\textbf{Difficulty Level:} \{difficulty\}

\textbf{Instructions:} Provide a comprehensive answer within \{word\_limit\} words. Ensure you cover all important aspects and key points related to the question. Be thorough but concise.

\textbf{Answer:}
\end{tcolorbox}

\section{Prompt Templates and Qualitative Examples}
\label{appendix:prompt}

This section provides detailed examples of the input prompts used in our evaluation, as well as qualitative examples demonstrating how model performance varies across the three settings (\textit{Base}, \textit{Mix}, and \textit{Golden}).

Crucially, regarding the Open-Ended (OE) prompts, we implemented a \textbf{Complexity-Adaptive Prompting} strategy to mitigate verbosity bias. The parameter \texttt{\{word\_limit\}} in the OE prompts is not fixed; it is dynamically adjusted based on the question's difficulty (i.e., the number of gold keypoints), enforcing concise answers for simple queries and allowing elaboration only for complex scenarios.

\subsection{Prompt Construction}
Prompt templates follow the general generation procedure outlined in Section~\ref{sec:evaluation_methodology}.
Table~\ref{tab:mcq_prompt} (end of section) illustrates the generation process from a raw passage to a structured MCQ.

\subsection{Qualitative Examples across Settings}
To intuitively understand the difference between the settings defined in Section~\ref{sec:eval_settings}, we provide concrete examples in Table~\ref{tab:mcq_base_mix_golden} (MCQ) and Table~\ref{tab:oe_base_mix_golden} (OE).
These examples illustrate:
\begin{itemize}
    \item \textbf{Base:} The model relies solely on internal parameters (often hallucinating).
    \item \textbf{Mix:} The model sees the gold passage alongside distractors (simulating noisy retrieval).
    \item \textbf{Golden:} The model sees only the correct passage (idealized context).
\end{itemize}

\paragraph{MCQ Generation Prompt.}
Table~\ref{tab:mcq_prompt} demonstrates how we used LLMs to generate the initial MCQ drafts from passages.

\section{Detailed Analysis by Disaster Type}
\label{appendix:event_breakdown}

To complement the aggregate results in the main text, we provide a fine-grained performance breakdown across the eight disaster categories defined in \textsc{DisastQA}.

\begin{figure}[h]
\centering
\includegraphics[width=0.5\textwidth]{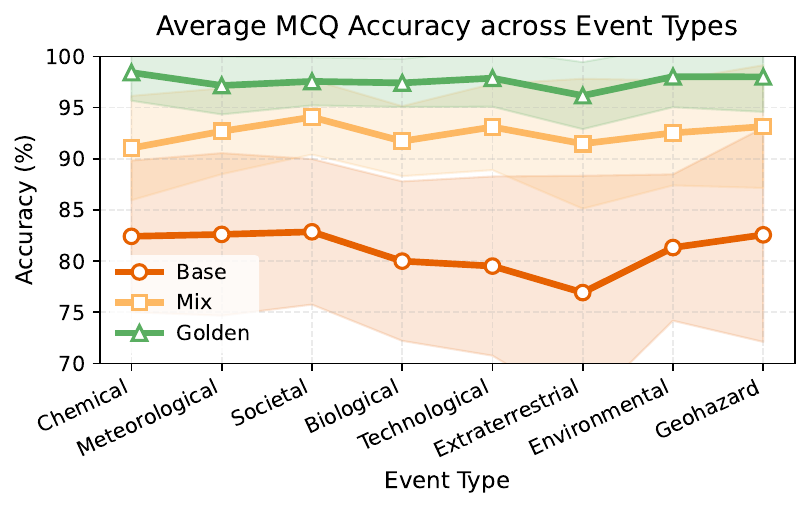}
\caption{
\textbf{MCQ Accuracy breakdown by Event Type.}
The gap between \textit{Base} and \textit{Golden} is most pronounced in specialized domains (e.g., Biological, Extraterrestrial), confirming that models rely on retrieval for long-tail knowledge.
}
\label{fig:event_breakdown_appendix}
\end{figure}

\begin{table*}[t]
\centering
\footnotesize
\setlength{\tabcolsep}{4pt}
\renewcommand{\arraystretch}{1.3}
\begin{tabular}{p{0.25\textwidth} p{0.15\textwidth} p{0.55\textwidth}}
\hline 
\textbf{Setting} & \textbf{Prediction} & \textbf{Retrieved Passage Content (Top-1)} \\

\hline 
\multicolumn{3}{l}{\textbf{MCQ Example 1: Volcanic Hazards}} \\
\hline 
\textbf{Question:} & \multicolumn{2}{p{0.70\textwidth}}{\textit{What model is utilized to enhance the accuracy of tephra dispersion predictions during volcanic eruptions?}} \\
\hline 
Base & \textbf{C (Wrong)} & \textit{No passage provided} \\
Mix & \textbf{C (Wrong)} & \textbf{[Distractor]} In: “Volcanic Hazards, Risks, and Disasters”, Eds: Shroder JF... This text discusses general hazards but mimics the terminology of dispersion models... \\
Golden & \textbf{B (Correct)} & \textbf{[Gold]} Predicting Tephra Dispersion with a Mesoscale Atmospheric Model and a Particle Fall Model: Application to Cerro Negro Volcano. The study utilizes a coupled model approach... \\
\hline 
\textbf{Correct Answer:} & \multicolumn{2}{l}{\textbf{B}} \\

\hline
\multicolumn{3}{l}{\textbf{MCQ Example 2: Shrimp Farming}} \\
\hline
\textbf{Question:} & \multicolumn{2}{p{0.70\textwidth}}{\textit{What recent advancement has been made in the fight against White Spot Virus in shrimp farming?}} \\
\hline
Base & \textbf{B (Wrong)} & \textit{No passage provided} \\
Mix & \textbf{D (Wrong)} & \textbf{[Distractor]} The company does not have the financial resources to follow this plan. In the meantime, Aquamen remains closed due to viral outbreaks... \\
Golden & \textbf{A (Correct)} & \textbf{[Gold]} Just saw a breakthrough in combating White Spot Virus! Researchers in India have developed a promising vaccine for shrimp that has shown high efficacy... \\
\hline
\textbf{Correct Answer:} & \multicolumn{2}{l}{\textbf{A}} \\

\hline
\multicolumn{3}{l}{\textbf{MCQ Example 3: Blowing Dust}} \\
\hline
\textbf{Question:} & \multicolumn{2}{p{0.70\textwidth}}{\textit{What is the horizontal visibility range characteristic of blowing dust according to the World Meteorological Organization?}} \\
\hline
Base & \textbf{C (Wrong)} & \textit{No passage provided} \\
Mix & \textbf{C (Wrong)} & \textbf{[Distractor]} It should be noted that due to the geographical and meteorological conditions (desert climate), El Paso may experience reduced visibility... \\
Golden & \textbf{D (Correct)} & \textbf{[Gold]} The World Meteorological Organization (WMO) categorizes dust events based on horizontal visibility: blowing dust is characterized by visibility reduction to between 1km and 10km... \\
\hline
\textbf{Correct Answer:} & \multicolumn{2}{l}{\textbf{D}} \\

\hline 
\end{tabular}
\caption{
Qualitative analysis of MCQ failure cases. 
In the \textbf{Mix} setting, models are often misled by \textbf{semantically relevant distractors} (marked as [Distractor]) that share lexical overlap with the query. 
Only the \textbf{Golden} setting enables accurate reasoning.
}
\label{tab:mcq_base_mix_golden}
\end{table*}

\clearpage

\begin{table*}[t]
\centering
\footnotesize
\setlength{\tabcolsep}{4pt}
\renewcommand{\arraystretch}{1.3}
\begin{tabular}{p{0.22\textwidth} p{0.45\textwidth} p{0.27\textwidth}}
\hline
\textbf{Setting} & \textbf{Model Response \& Coverage} & \textbf{Context Availability} \\

\hline
\multicolumn{3}{l}{\textbf{OE Example 1: Laboratory Ventilation}} \\
\hline
\textbf{Question:} & \multicolumn{2}{p{0.72\textwidth}}{\textit{What engineering controls and operational strategies should be implemented in laboratory ventilation systems to effectively prevent fume backflow...?}} \\
\hline
Base & To effectively prevent fume backflow... ensure safety... [General knowledge, vague details] \newline \textbf{(0\% Coverage - Poor)} & \textit{No passage provided} \\
\hline
Mix & Passage 5 states that to effectively prevent fume backflow... [Partial hallucination mixed with retrieval] \newline \textbf{(75\% Coverage - Partial)} & \textbf{[Gold + 4 Distractors]} \newline \textit{Model struggles to isolate Gold details.} \\
\hline
Golden & In laboratory environments, effective ventilation systems... [Correctly cites engineering controls] \newline \textbf{(100\% Coverage - Good)} & \textbf{[Gold Only]} \newline In effect, this means arranging... \\
\hline
\textbf{Reference:} & \multicolumn{2}{p{0.72\textwidth}}{The passage outlines several engineering controls and operational strategies for laboratory ventilation systems to prevent fume backflow and ensure safety and comfort for personnel and animals.} \\

\hline
\multicolumn{3}{l}{\textbf{OE Example 2: Waste Management}} \\
\hline
\textbf{Question:} & \multicolumn{2}{p{0.72\textwidth}}{\textit{What specific strategies can be implemented in public works projects to enhance waste management practices in the construction and demolition sector...?}} \\
\hline
Base & To enhance waste management... [Specific strategies focus on...] \newline \textbf{(0\% Coverage - Poor)} & \textit{No passage provided} \\
\hline
Mix & Passage 4 suggests that to enhance waste management practices... [Retrieval noise leads to omission] \newline \textbf{(67\% Coverage - Partial)} & \textbf{[Gold + 4 Distractors]} \newline \textit{Retrieval noise leads to omission.} \\
\hline
Golden & To enhance waste management practices... [Includes targeted strategies] \newline \textbf{(100\% Coverage - Good)} & \textbf{[Gold Only]} \newline This can be achieved by... \\
\hline
\textbf{Reference:} & \multicolumn{2}{p{0.72\textwidth}}{The passage outlines several strategies that can be implemented in public works projects to enhance waste management practices in the construction and demolition sector.} \\

\hline
\multicolumn{3}{l}{\textbf{OE Example 3: IoT in Crowd Stampedes}} \\
\hline
\textbf{Question:} & \multicolumn{2}{p{0.72\textwidth}}{\textit{How are Internet of Things (IoT) technologies being utilized in risk assessment and management strategies for early-warning systems designed to prevent crowd stampedes...?}} \\
\hline
Base & Internet of Things (IoT) technologies are increasingly being integrated... \newline \textbf{(33\% Coverage - Poor)} & \textit{No passage provided} \\
\hline
Mix & Passage 2 mentions that IoT technologies... \newline \textbf{(75\% Coverage - Partial)} & \textbf{[Gold + 4 Distractors]} \newline \textit{Partially correct but mixes sources.} \\
\hline
Golden & Internet of Things (IoT) technologies are increasingly being leveraged... \newline \textbf{(80\% Coverage - Good)} & \textbf{[Gold Only]} \newline This paper discusses the analysis... \\
\hline
\textbf{Reference:} & \multicolumn{2}{p{0.72\textwidth}}{The passage describes the utilization of Internet of Things (IoT) technologies in risk assessment and management strategies for early-warning systems aimed at preventing crowd stampedes.} \\

\hline
\end{tabular}
\caption{
Progression of Open-Ended (OE) response quality. 
\textbf{Base} models generate fluent but factually empty responses. 
\textbf{Mix} settings improve coverage but often include irrelevant details from noise. 
\textbf{Golden} context yields the most comprehensive Keypoint Coverage.
}
\label{tab:oe_base_mix_golden}
\end{table*}

\begin{table*}[t]
\centering
\small

\resizebox{\textwidth}{!}{
\begin{tabular}{lcccc|lcccc}
\toprule
\multicolumn{5}{c|}{\textbf{Part 1 Models}} & \multicolumn{5}{c}{\textbf{Part 2 Models}} \\
\cmidrule(r){1-5} \cmidrule(l){6-10}
\textbf{Difficulty} & \textbf{Base} & \textbf{Mix} & \textbf{Golden} & \textbf{Avg} &
\textbf{Difficulty} & \textbf{Base} & \textbf{Mix} & \textbf{Golden} & \textbf{Avg} \\
\midrule
\multicolumn{5}{l|}{\textbf{AceMath-1.5B-Instr.}} & \multicolumn{5}{l}{\textbf{Llama-3-8B}} \\
Easy & 71.76 & 85.17 & \textbf{85.99} & 80.97 & Easy & 74.99 & 84.52 & \textbf{96.28} & 85.26 \\
Medium & 68.44 & 85.36 & \textbf{92.36} & 82.05 & Medium & 75.31 & 88.75 & \textbf{91.90} & 85.32 \\
Hard & 63.87 & 88.52 & \textbf{94.13} & 82.17 & Hard & 82.21 & 90.09 & \textbf{91.02} & 87.77 \\
Extremely Complex & 80.62 & 83.75 & \textbf{90.18} & 84.85 & Extremely Complex & 85.42 & \textbf{100.00} & 83.03 & 89.48 \\
\midrule
\multicolumn{5}{l|}{\textbf{DeepSeek-v3-7B}} & \multicolumn{5}{l}{\textbf{Mistral-7B-Instr.}} \\
Easy & 72.33 & \textbf{88.61} & 88.60 & 83.18 & Easy & 76.00 & 85.34 & \textbf{88.88} & 83.41 \\
Medium & 71.55 & 86.80 & \textbf{90.34} & 82.90 & Medium & 75.67 & 84.59 & \textbf{89.99} & 83.42 \\
Hard & 81.37 & 85.62 & \textbf{88.55} & 85.18 & Hard & 78.82 & 81.28 & \textbf{88.27} & 82.79 \\
Extremely Complex & 90.00 & 85.42 & \textbf{100.00} & 91.81 & Extremely Complex & 82.50 & 77.08 & \textbf{90.83} & 83.47 \\
\midrule
\multicolumn{5}{l|}{\textbf{Falcon-3-1B-Instr.}} & \multicolumn{5}{l}{\textbf{Phi-2}} \\
Easy & 77.32 & 87.33 & \textbf{91.31} & 85.32 & Easy & 75.80 & 81.10 & \textbf{87.27} & 81.39 \\
Medium & 71.90 & 84.03 & \textbf{92.61} & 82.85 & Medium & 69.63 & 81.87 & \textbf{89.78} & 80.43 \\
Hard & 74.71 & 85.29 & \textbf{91.69} & 83.90 & Hard & 76.46 & 83.46 & \textbf{86.67} & 82.20 \\
Extremely Complex & 59.58 & 83.75 & \textbf{92.86} & 78.73 & Extremely Complex & 68.75 & 80.18 & \textbf{88.75} & 79.23 \\
\midrule
\multicolumn{5}{l|}{\textbf{Gemini-1.5 Pro}} & \multicolumn{5}{l}{\textbf{Qwen-2.5-3B-Instr.}} \\
Easy & 79.96 & 91.17 & \textbf{95.44} & 88.86 & Easy & 74.98 & 85.71 & \textbf{96.17} & 85.62 \\
Medium & 82.57 & 88.26 & \textbf{94.15} & 88.33 & Medium & 76.89 & 87.54 & \textbf{91.98} & 85.47 \\
Hard & 83.96 & 90.40 & \textbf{95.38} & 89.91 & Hard & 80.33 & 87.50 & \textbf{92.71} & 86.85 \\
Extremely Complex & 86.67 & 81.66 & \textbf{88.75} & 85.69 & Extremely Complex & 81.67 & \textbf{90.83} & 81.43 & 84.64 \\
\midrule
\multicolumn{5}{l|}{\textbf{Gemma-7B}} & \multicolumn{5}{l}{\textbf{Qwen-3-0.6B}} \\
Easy & 66.85 & 86.36 & \textbf{93.67} & 82.29 & Easy & 65.02 & 77.59 & \textbf{93.19} & 78.60 \\
Medium & 75.23 & 83.59 & \textbf{91.53} & 83.45 & Medium & 71.59 & 86.24 & \textbf{91.05} & 82.96 \\
Hard & 75.68 & 85.32 & \textbf{93.41} & 84.80 & Hard & 79.91 & 85.96 & \textbf{91.85} & 85.91 \\
Extremely Complex & 75.42 & 77.14 & \textbf{90.83} & 81.13 & Extremely Complex & 59.38 & \textbf{91.67} & \textbf{91.67} & 80.91 \\
\midrule
\multicolumn{5}{l|}{\textbf{GPT-4o}} & \multicolumn{5}{l}{\textbf{Qwen-3-4B}} \\
Easy & 82.89 & \textbf{96.22} & 96.06 & 91.72 & Easy & 79.93 & 86.56 & \textbf{90.56} & 85.68 \\
Medium & 84.42 & 91.87 & \textbf{95.25} & 90.51 & Medium & 77.42 & 87.35 & \textbf{93.11} & 85.96 \\
Hard & 80.07 & 93.06 & \textbf{95.56} & 89.56 & Hard & 78.05 & \textbf{86.22} & 83.75 & 82.67 \\
Extremely Complex & 89.58 & 90.83 & \textbf{100.00} & 93.47 & Extremely Complex & 85.83 & \textbf{93.75} & 86.66 & 88.75 \\
\midrule
\multicolumn{5}{l|}{\textbf{Hunyuan-4B-Instr.}} & \multicolumn{5}{l}{\textbf{Qwen-3-8B}} \\
Easy & 58.79 & 85.47 & \textbf{92.50} & 78.92 & Easy & 76.35 & 87.26 & \textbf{94.17} & 85.93 \\
Medium & 66.49 & 86.13 & \textbf{89.64} & 80.75 & Medium & 78.37 & 89.64 & \textbf{94.46} & 87.49 \\
Hard & 71.81 & 90.33 & \textbf{92.63} & 84.92 & Hard & 76.50 & \textbf{93.80} & 93.62 & 87.97 \\
Extremely Complex & \textbf{95.83} & 86.66 & 81.61 & 88.03 & Extremely Complex & 65.00 & 74.17 & \textbf{87.50} & 75.56 \\
\midrule
\multicolumn{5}{l|}{\textbf{Hunyuan-7B-Instruct}} & \multicolumn{5}{l}{\textbf{Yi-6B-Chat}} \\
Easy & 36.75 & \textbf{84.74} & 76.39 & 65.96 & Easy & 67.48 & 85.71 & \textbf{86.50} & 79.90 \\
Medium & 39.90 & \textbf{86.45} & 79.36 & 68.57 & Medium & 71.72 & 84.87 & \textbf{86.23} & 80.94 \\
Hard & 53.02 & \textbf{87.88} & 78.22 & 73.04 & Hard & 76.85 & 83.04 & \textbf{88.65} & 82.85 \\
Extremely Complex & 58.34 & \textbf{80.42} & 56.25 & 65.00 & Extremely Complex & 95.83 & 96.43 & \textbf{100.00} & 97.42 \\
\midrule
\multicolumn{5}{l|}{\textbf{Llama-3.2-1B-Instr.}} & \multicolumn{5}{l}{\textbf{GPT-5.2}} \\
Easy & 82.10 & 77.28 & \textbf{93.66} & 84.35 & Easy & \textbf{91.03} & 94.66 & 89.22 & 91.64 \\
Medium & 72.27 & 82.01 & \textbf{90.00} & 81.43 & Medium & 86.20 & 92.33 & \textbf{95.51} & 91.35 \\
Hard & 72.47 & 87.16 & \textbf{89.63} & 83.09 & Hard & 85.89 & \textbf{96.05} & 95.23 & 92.39 \\
Extremely Complex & 75.59 & 87.50 & \textbf{95.00} & 86.03 & Extremely Complex & \textbf{100.00} & 95.00 & \textbf{100.00} & 98.33 \\
\midrule
\multicolumn{5}{l|}{\textbf{Llama-3.2-3B-Instr.}} & \multicolumn{5}{l}{\textbf{Gemini-3 Pro}} \\
Easy & 66.18 & 86.25 & \textbf{89.37} & 80.60 & Easy & 83.13 & 89.72 & \textbf{94.36} & 89.07 \\
Medium & 76.41 & 85.12 & \textbf{94.04} & 85.19 & Medium & 85.01 & 92.59 & \textbf{97.07} & 91.56 \\
Hard & 77.02 & 87.62 & \textbf{88.89} & 84.51 & Hard & 85.29 & 91.12 & \textbf{96.18} & 90.86 \\
Extremely Complex & 88.75 & \textbf{95.83} & 95.00 & 93.19 & Extremely Complex & 78.87 & 75.00 & \textbf{100.00} & 84.62 \\
\bottomrule
\end{tabular}}
\caption{
Coverage (\%) across difficulty levels and retrieval settings (Base, Mix, Golden) 
for all 20 models. Each column group corresponds to one model family. 
The best value within each difficulty row is highlighted in bold.
}
\label{tab:coverage_difficulty_models_all}
\end{table*}

\begin{table*}[t]
\centering
\footnotesize
\setlength{\tabcolsep}{2pt}
\renewcommand{\arraystretch}{0.95}

\scalebox{0.95}{\resizebox{\textwidth}{!}{%
\tiny
\begin{tabular}{lcccc}
\toprule
\textbf{Model / Event} &
\textbf{Chemical} & 
\textbf{Meteorological} & 
\textbf{Societal} & 
\textbf{Biological} \\
\midrule
\multicolumn{1}{l}{\textbf{Closed-source Models}}
& base / mix / golden
& base / mix / golden
& base / mix / golden
& base / mix / golden
\\[0.5pt]
GPT-5.2                   & 95.2 / 95.2 / 99.6 & 92.4 / 96.4 / 99.2 & 94.8 / 97.2 / 99.6 & 90.8 / 95.6 / 100.0 \\
Gemini-3 Pro               & 93.6 / 94.4 / 95.2 & 92.4 / 94.0 / 95.6 & 93.6 / 96.0 / 97.2 & 90.8 / 96.8 / 97.6 \\
GPT-4o                    & 93.6 / 94.0 / 99.6 & 88.8 / 96.4 / 98.8 & 89.2 / 96.4 / 99.6 & 91.6 / 94.8 / 98.4 \\
Gemini-1.5 Pro            & 89.6 / 88.2 / 98.8 & 90.8 / 90.2 / 98.4 & 91.6 / 94.4 / 98.0 & 89.2 / 92.3 / 99.3 \\
\midrule
\multicolumn{5}{l}{\textbf{Large Open Models (6B--8B)}} \\[0.5pt]
Qwen-3-8B                 & 88.0 / 93.6 / 100.0 & 89.2 / 98.0 / 100.0 & 92.4 / 97.2 / 99.6 & 85.2 / 95.2 / 99.6 \\
Llama-3-8B                & 86.4 / 93.6 / 100.0 & 84.4 / 95.2 / 98.0 & 87.6 / 96.4 / 98.8 & 85.6 / 95.6 / 99.2 \\
Yi-6B-Chat                & 87.6 / 94.4 / 100.0 & 86.4 / 94.8 / 98.8 & 88.4 / 98.0 / 99.6 & 82.0 / 95.2 / 98.8 \\
DeepSeek-v3-7B            & 81.2 / 95.2 / 100.0 & 82.4 / 94.4 / 98.4 & 85.6 / 95.6 / 98.4 & 78.8 / 92.4 / 97.6 \\
Gemma-7B                  & 81.2 / 86.8 / 98.4 & 85.6 / 90.8 / 98.0 & 82.4 / 92.4 / 98.4 & 78.8 / 88.8 / 97.2 \\
Mistral-7B-Instruct-v0.2  & 80.0 / 92.4 / 99.6 & 84.8 / 95.6 / 99.2 & 82.4 / 95.6 / 98.4 & 78.4 / 92.4 / 98.4 \\
Hunyuan-7B-Instruct       & 64.4 / 86.8 / 93.6 & 56.8 / 87.2 / 90.4 & 64.8 / 89.2 / 93.2 & 54.0 / 88.0 / 94.8 \\
\midrule
\multicolumn{5}{l}{\textbf{Medium Open Models (3B--5B)}} \\[0.5pt]
Qwen-3-4B                 & 85.6 / 93.2 / 100.0 & 88.4 / 96.8 / 99.2 & 87.6 / 96.4 / 98.8 & 85.6 / 94.8 / 99.6 \\
Qwen-2.5-3B-Instruct      & 89.2 / 95.6 / 100.0 & 87.2 / 96.0 / 99.6 & 87.6 / 96.8 / 98.4 & 81.2 / 93.6 / 98.4 \\
Llama-3.2-3B-Instruct     & 86.8 / 94.8 / 99.6 & 84.8 / 95.6 / 98.8 & 83.6 / 94.8 / 97.6 & 84.0 / 94.0 / 98.8 \\
Hunyuan-4B-Instruct       & 83.6 / 93.6 / 99.2 & 85.2 / 93.2 / 96.4 & 80.4 / 95.2 / 96.8 & 81.2 / 91.6 / 98.0 \\
Phi-2                     & 87.6 / 95.6 / 100.0 & 89.6 / 94.8 / 96.8 & 88.4 / 96.0 / 98.4 & 85.6 / 94.0 / 99.2 \\
\midrule
\multicolumn{5}{l}{\textbf{Small Open Models ($\leq$2B)}} \\[0.5pt]
AceMath-1.5B-Instruct     & 77.6 / 92.8 / 99.6 & 75.6 / 90.8 / 97.2 & 72.0 / 94.0 / 98.8 & 76.0 / 91.6 / 97.2 \\
Falcon-3-1B-Instruct      & 77.6 / 87.2 / 98.0 & 79.6 / 92.0 / 95.6 & 75.2 / 92.0 / 97.2 & 73.6 / 89.2 / 96.4 \\
Llama-3.2-1B-Instruct     & 81.6 / 92.4 / 98.4 & 83.6 / 92.8 / 98.0 & 83.6 / 94.8 / 97.2 & 79.2 / 91.2 / 95.2 \\
Qwen-3-0.6B               & 79.2 / 85.2 / 88.8 & 72.8 / 84.8 / 89.2 & 74.0 / 84.8 / 89.6 & 74.4 / 84.4 / 89.6 \\
\bottomrule
\end{tabular}%
}}

\vspace{2pt}
\noindent\emph{Note: Values are \textbf{Base / Mix / Golden}. Top panel shows the first four event types (Chemical, Meteorological, Societal, Biological). Bottom panel shows the remaining four (Technological, Extraterrestrial, Environmental, Geohazard).}
\vspace{3pt}

\scalebox{0.95}{\resizebox{\textwidth}{!}{%
\tiny
\begin{tabular}{lcccc}
\toprule
\textbf{Model / Event} &
\textbf{Technological} &
\textbf{Extraterrestrial} &
\textbf{Environmental} &
\textbf{Geohazard} \\
\midrule
\multicolumn{1}{l}{\textbf{Closed-source Models}}
& base / mix / golden
& base / mix / golden
& base / mix / golden
& base / mix / golden
\\[0.5pt]
GPT-5.2                   & 92.4 / 96.4 / 99.6 & 89.6 / 98.0 / 99.2 & 96.0 / 96.4 / 100.0 & 94.0 / 98.4 / 100.0 \\
Gemini-3 Pro               & 94.4 / 95.2 / 96.4 & 87.2 / 94.4 / 97.2 & 93.6 / 95.2 / 97.6 & 92.0 / 96.8 / 96.8 \\
GPT-4o                    & 90.0 / 96.8 / 100.0 & 90.8 / 98.0 / 99.2 & 90.0 / 95.6 / 99.2 & 94.4 / 98.0 / 100.0 \\
Gemini-1.5 Pro            & 89.2 / 91.7 / 99.6 & 83.6 / 90.7 / 96.8 & 88.0 / 86.8 / 99.6 & 89.6 / 92.9 / 99.2 \\
\midrule
\multicolumn{5}{l}{\textbf{Large Open Models (6B--8B)}} \\[0.5pt]
Qwen-3-8B                 & 89.2 / 97.2 / 100.0 & 84.8 / 96.4 / 99.6 & 88.0 / 96.0 / 100.0 & 92.4 / 96.4 / 100.0 \\
Llama-3-8B                & 86.4 / 96.4 / 100.0 & 87.2 / 94.8 / 97.6 & 88.4 / 96.0 / 99.6 & 91.2 / 97.2 / 100.0 \\
Yi-6B-Chat                & 85.2 / 95.6 / 99.2 & 83.6 / 96.0 / 98.4 & 83.2 / 94.4 / 100.0 & 89.6 / 96.8 / 99.6 \\
DeepSeek-v3-7B            & 79.6 / 94.0 / 99.2 & 72.0 / 92.8 / 97.6 & 80.0 / 94.8 / 99.2 & 82.0 / 95.6 / 98.8 \\
Gemma-7B                  & 80.0 / 91.2 / 98.0 & 71.6 / 87.2 / 97.2 & 83.6 / 88.8 / 98.4 & 78.8 / 92.0 / 98.8 \\
Mistral-7B-Instruct-v0.2  & 80.0 / 95.6 / 99.2 & 82.4 / 93.2 / 97.2 & 80.0 / 95.2 / 98.0 & 86.4 / 94.8 / 99.2 \\
Hunyuan-7B-Instruct       & 54.0 / 86.8 / 93.2 & 40.4 / 86.0 / 90.0 & 65.6 / 89.2 / 94.4 & 56.0 / 90.0 / 94.8 \\
\midrule
\multicolumn{5}{l}{\textbf{Medium Open Models (3B--5B)}} \\[0.5pt]
Qwen-3-4B                 & 86.4 / 96.8 / 100.0 & 81.2 / 96.0 / 99.6 & 84.4 / 96.4 / 99.6 & 89.2 / 97.6 / 100.0 \\
Qwen-2.5-3B-Instruct      & 84.4 / 96.0 / 99.6 & 81.2 / 94.8 / 98.0 & 81.6 / 94.8 / 100.0 & 90.0 / 96.0 / 99.2 \\
Llama-3.2-3B-Instruct     & 81.2 / 95.2 / 98.4 & 86.4 / 96.0 / 97.6 & 84.8 / 95.2 / 99.2 & 89.2 / 96.8 / 99.2 \\
Hunyuan-4B-Instruct       & 74.8 / 94.4 / 97.2 & 82.8 / 94.8 / 96.8 & 82.8 / 95.2 / 98.4 & 84.8 / 96.8 / 99.6 \\
Phi-2                     & 86.8 / 97.2 / 99.6 & 83.6 / 95.2 / 98.4 & 89.2 / 96.0 / 99.2 & 87.2 / 96.8 / 100.0 \\
\midrule
\multicolumn{5}{l}{\textbf{Small Open Models ($\leq$2B)}} \\[0.5pt]
AceMath-1.5B-Instruct     & 69.2 / 92.4 / 99.6 & 65.6 / 89.6 / 94.8 & 79.2 / 93.6 / 99.6 & 77.6 / 92.0 / 98.4 \\
Falcon-3-1B-Instruct      & 74.0 / 92.4 / 96.0 & 70.8 / 91.6 / 96.4 & 74.4 / 94.0 / 98.8 & 72.8 / 91.2 / 96.8 \\
Llama-3.2-1B-Instruct     & 78.0 / 92.4 / 95.2 & 76.8 / 94.0 / 95.6 & 84.8 / 95.6 / 99.2 & 89.2 / 96.8 / 99.2 \\
Qwen-3-0.6B               & 70.8 / 82.4 / 88.8 & 75.2 / 77.6 / 86.4 & 72.0 / 84.0 / 87.6 & 78.0 / 80.0 / 84.8 \\
\bottomrule
\end{tabular}%
}}

\caption{
MCQ accuracy (\%) under \textbf{Base / Mix / Golden} across all eight \textbf{Event Types}, shown in two panels for readability.  
\textbf{Top:} Chemical, Meteorological, Societal, Biological.  
\textbf{Bottom:} Technological, Extraterrestrial, Environmental, Geohazard.  
Models are grouped by access type and parameter scale.
}
\label{tab:mcq_event_all}
\end{table*}

\end{document}